\documentclass{article}

    \title{OFTER: An online pipeline for time series forecasting}
\usepackage[list=true]{subcaption}

\usepackage{arxiv}
\usepackage[utf8]{inputenc} % allow utf-8 input
\usepackage{url}            % simple URL typesetting
\usepackage{booktabs}       % professional-quality tables
\usepackage{amsfonts}       % blackboard math symbols
\usepackage{nicefrac}       % compact symbols for 1/2, etc.
\usepackage{microtype}      % microtypography
\usepackage{lipsum}
\usepackage{amsmath}
\usepackage{bbm}
\usepackage{float}
\usepackage{mathtools}  
\usepackage{amssymb}
\usepackage{tabulary}
\usepackage{amsthm}
\usepackage[titletoc]{appendix}
\usepackage[pdftex]{graphicx}

\usepackage[export]{adjustbox}
\usepackage{graphicx}  
\usepackage{chngcntr}
\usepackage{pdfpages}
\usepackage{lineno}
\DeclarePairedDelimiterX\set[1]\lbrace\rbrace{#1}

\usepackage{xcolor}
\usepackage{authblk}
\usepackage[colorlinks=true,linkcolor=red,filecolor=green,citecolor=blue]{hyperref}
\usepackage{cleveref}

\usepackage{tcolorbox}
\usepackage{placeins}
\usepackage{array,xcolor,colortbl}
\usepackage{multirow}

\newcolumntype{G}{>{\centering\columncolor{gray!20!white}}p{0.2\textwidth}}
\newcolumntype{C}{>{\centering\arraybackslash}p{0.2\textwidth}}
\usepackage{verbatim}

\usepackage{titletoc}
    \titlecontents{chapter}
    [0em] %
    {\medskip\bfseries}
    {\thecontentslabel\quad}%numbered chapters
    {}%numberless chapter
    {\hfill\contentspage}

    \titlecontents{section}[1.72em]{\smallskip}%
    {\thecontentslabel.\enspace}%numbered sections
    {}%numberless section
    {\titlerule*[1.2pc]{.}\contentspage}%

\definecolor{colour3}{RGB}{178,55,250} % purple

\newcounter{noteMCctr} 

\newcounter{noteSHctr} 

\newcounter{noteNMctr} 

\newcommand{\bb}{\hspace{0mm}$\bullet$}
\usepackage[linesnumbered,ruled,vlined]{algorithm2e}
\usepackage{algorithmic}

\author{Nikolas Michael$^{\dag}$\thanks{Corresponding author.
Email: nikolas.michael@kellogg.ox.ac.uk} , Mihai Cucuringu$^{\dag \ddag \odot \S}$ and Sam Howison$^{\ddag}$} 

\affil{
$\dag$Department of Statistics, University of Oxford, 24-29 St Giles', Oxford OX1 3LB, UK\\
$\ddag$Mathematical Institute, University of Oxford, Andrew Wiles Building, Woodstock Rd, Oxford OX2 6GG \\
$\S$  Oxford-Man Institute of Quantitative Finance, University of Oxford, Oxford, UK \\ 
$\odot$ The Alan Turing Institute, John Dodson House, 96 Euston Rd, London NW1 2DB
}

\date{April 2023}

 \let\originalparagraph\paragraph
    \renewcommand{\paragraph}[2][.]{\originalparagraph{#2#1}}

\crefname{algocf}{alg.}{algs.}
\Crefname{algorithm}{Algorithm}{Algorithms}
\Crefname{Algorithm}{Algorithm}{Algorithms}
\Crefname{algorithm}{algorithm}{algorithm}

\DeclarePairedDelimiter\floor{\lfloor}{\rfloor}
\newcommand{\ofter}{\textsc{OFTER}}

\begin{document}

\maketitle

\begin{abstract}
We introduce {\ofter}, a time series forecasting pipeline tailored for mid-sized multivariate time series. {\ofter} utilizes the non-parametric models of k-nearest neighbors and Generalized Regression Neural Networks, integrated with a dimensionality reduction component. To circumvent the curse of dimensionality, we employ a weighted norm based on a modified version of the maximal correlation coefficient. The pipeline we introduce is specifically designed for online tasks, has an interpretable output, and is able to outperform several  state-of-the art baselines. The computational efficacy of the algorithm, its online nature, and its ability to operate in low signal-to-noise regimes, render {\ofter} an ideal approach for financial multivariate time series problems, such as daily equity forecasting. Our work demonstrates that while deep learning models hold significant promise for time series forecasting, traditional methods carefully integrating mainstream tools remain very competitive alternatives with the added benefits of scalability and interpretability.
\end{abstract}


\hypersetup{linkcolor=blue}
\tableofcontents

\section{Introduction} \label{seq:intro}
Time series are temporally structured data sets that appear in most disciplines, ranging from physics to the social sciences. For forecasting tasks, the temporal structure can be leveraged to improve performance but also to avoid look-ahead biases in the training phase. For univariate time series, parametric linear time series models such as the Autoregressive Integrated Moving Average (ARIMA) and the Autoregressive Conditional Heteroscedasticity (ARCH) have attained sufficiently good results in most applications of interest.
While these algorithms can be extended to the setting of multivariate time series, the rapid rise in computing power in recent decades has rendered machine learning techniques an attractive alternative to traditional time series methods. An early comparison between different machine learning  models  was \cite{nerseen}, where the authors concluded that the choice of method had a significant effect in performance.

Deep learning models have been a popular choice for multivariate time series forecasting, including Long-Short Term Memory (LSTM) \cite{lstm}, Bidirectional LSTM \cite{bilstm} and Gated Recursion Unit \cite{gru} (GRU). Various expansions of such deep learning models have been proposed in recent years, such as DeepAR \cite{deepar} and Temporal Fusion Transformers \cite{tftemporal}, which generally achieve state-of-the-art performances across various domains.
However, such approaches often employ thousands of trainable parameters, which can be difficult to train with a limited number of observations, thus leading to degraded out-of-sample forecasting performance.
Despite such advancements in deep learning, simple regression-based models are still widely used and sought after. This is especially true for mid-low frequency financial tasks, such predicting daily returns, volumes, and realized volatilities, partly due to the complexity, unpredictability and constantly shifting dynamics of the financial markets. For example, Ordinary Least Squares (OLS) is still widely used for both forecasting and interpretive statistical analysis due to its simplicity, closed-form solution, and the fact that it requires no choice of hyper-parameters. Its regularized versions, such as the LASSO and the Elastic Net are some of the most widely utilized regression models in finance, as they require only a few hyperparameters.

On the other hand, various recent lines of work have demonstrated the merit of combining traditional time series models and machine learning models, such as \cite{Domingos, FIRMINO, PANIGRAHI,PAI,ZHANG_hybrid}. In this paper, we consider one of the simplest types of hybrid models, which builds on the results of a univariate ARIMA model and enhances it by performing a novel regression method on the residuals of an ARIMA model.

\paragraph{Non-parametric models and dimensionality reduction} 
Two simple models that have been relatively overlooked by the machine learning community, are k-Nearest neighbors (kNN) and Generalized Regression Neural Networks (GRNN), both of which are simple, non-parametric, fast methods requiring only a single hyperparameter. This mainly stems from the curse of dimensionality, which affects both techniques. GRNNs have been utilized primarily in engineering applications, such as \cite{zyga} for logo recognition, \cite{CROSS2018124} for forecasting feeding patterns, and \cite{pal2011modelling} for modeling the pile-bearing capacity of soil. An overview of its different applications for dynamical systems has been pursued by \cite{al-maha}. In this work, we aim to circumvent the curse of dimensionality that these two methods suffer from by using the following two techniques. Firstly, instead of the Euclidean norm employed by kNN and GRNN, we consider a thresholded weighted norm that is driven by the feature's predictive potential. Secondly, before employing these well-known models, we consider a dimensionality reduction component, which is by far the most common remedy for the curse of dimensionality. 

Dimensionality reduction is one of the main unsupervised machine learning paradigms, with the past couple of decades having witnessed seminal contributions in terms of non-linear dimensionality reduction methods, such as locally linear embedding (LLE) \cite{LLE}, ISOMAP \cite{TenenbaumEtAl2000}, Hessian LLE \cite{donoho2003HessianEigenmaps}, Laplacian eigenmaps \cite{BelkinLaplacianEigenmaps}, and diffusion maps \cite{Coifman2006DiffusionMaps}. In more recent years, deep learning models have also been increasingly used as a means of dimensionality reduction, such as the class of Autoencoders \cite{kramer}, which can be seen as a non-linear extension of PCA. Performing dimensionality reduction before employing a kNN technique has shown a significant improvement in performance \cite{pulgar2018aeknn}. Partial Least Squares is a popular method of incorporating information from the target into the initial features, during the embedding stage. More recently, \cite{spca} achieved improvements in the performance of PCA by scaling each factor based on its predictive slope with the target.

\paragraph{Uses in financial applications}
Examples of dimension reduction use cases in time series forecasting tasks include \cite{chiavazzo}, which used a truncated diffusion map to construct a global manifold for the slow dynamics of a reduced kinetic model, and \cite{papaioannou2021time} which utilizes manifold learning techniques in conjunction with various supervised learning models to build a profitable strategy for the Forex market. Along the same lines, \cite{YAN2022102521} proposed a hybrid approach integrating the Midas-RV model and PCA to achieve better results in forecasting the volatility of crude oil futures in the Chinese market.
One financial application we consider in this paper is the forecasting of daily traded volumes. Being able to accurately predict next-day volumes is of interest to practitioners and especially to market makers. Previous works include \cite{JOSEPH20111116}, who showed how search engine data can be leveraged to forecast future trading volumes. More similarly to our work is \cite{Liu2017IntradayVP}, who utilizes Support Vector Machines to predict the daily volume percentages with promising performance, in the spirit of         \cite{libman} who combine SVM and LSTM after residualizing with an AR model. 

\paragraph{Main contribution: {\ofter} - a novel integrated pipeline}
We consider the task of one-step time series forecasting and propose an approach that integrates ARIMA models, dimensionality reduction, and simple non-parametric methods into a single pipeline, which we denote as \textit{Online Forecasting via Temporal Embedding Regression ({\ofter})}. To alleviate the curse of high dimensionality and enhance our dimensionality reduction component, we conduct feature selection on the embedding features based on the concept of \textit{maximal correlation}. We apply {\ofter} to the financial tasks of forecasting next-day volume and returns by building an embedding based on various features of 10 Exchange Traded Funds (ETFs). Through an extensive set of numerical simulations on both synthetic and real data, we demonstrate that {\ofter} provides a powerful alternative to other mainstream machine learning models mid-sized data sets, even performing better than established time series-tailored deep learning models. In particular, it can work effectively in low signal-to-noise regimes for mid-sized data sets, while still being computationally efficient.

Furthermore, our proposed methodology provides the following additional benefits.
\begin{itemize} 
 \item Once an embedding has been computed, it is straightforward for the algorithm to perform out-of-sample extension and incorporate newly arrived time observations; therefore, {\ofter} can naturally be construed as an online algorithm that operates efficiently in an online setting.
\item Certain configurations of our method fall under the unsupervised learning setting, as the embedding is independent of the target. In other words, the same embedding can potentially be employed to forecast different target variables of interest, and thus {\ofter} is amenable to extensions to the multi-target setting.
\item Our method can further provide an avenue for change-point detection and anomaly detection tasks, which can be seen as natural byproducts. A previous or newly arrived observation without any local neighboring points in the embedding, could indicate either an extreme measurement or a change-point in the time series.
\end{itemize}

\paragraph{Paper Structure}
The rest of this paper is structured as follows. In \Cref{sec:background} we give an overview of some of the essential concepts used in the proposed pipeline, including the kNN and GRNN algorithms, principal component analysis, and maximal correlation. In \Cref{sec:main}, we introduce our proposed methodology, along with its main components and their integration into an algorithm. 
In \Cref{sec:results} we analyze the performance of {\ofter} in both synthetic experiments and real-world data. We also discuss the additional properties of {\ofter} in \Cref{sec:benefits}. Finally, we conclude in \Cref{sec:conclusion}. 

\section{Preliminaries and related works} \label{sec:background}
Consider a $d$-dimensional time series represented as a matrix $X = \left( x_1,\dots,x_T \right) \in \mathbb{R}^{T \times d}$ with rows denoting observations measures along $T$ time steps, which we refer to the feature matrix. Assume for the rest of this paper that we are interested in forecasting a single univariate target variable, which we denote by $y = \left( y_1,\dots,y_T \right) \in \mathbb{R}^{T}$ and refer to as the target variable. Furthermore, denote by $x' \in \mathbb{R}^{d}$ and $y' \in \mathbb{R}$, the value of the features and the target variable, respectively, associated with a new observation. We denote by $e_d$ the all-ones vector of length $d$. In terms of matrix indexing, we generally denote the (i,j) entry of $X$ by $X_{i,j}$, the submatrix of $X$ including rows starting from $t_1$ up to $t_2$ by $X_{t_1:t_2}$ , i.e $\left( x_{t_1},\dots,x_{t_2} \right)$ and by $X_{t_1:t_2,j}$ the $j^{th}$ column of $X_{t_1:t_2}$.

\subsection{K-nearest neighbors and Generalized Regression Neural Network} 
K-Nearest Neighbors (kNN) is a non-parametric supervised learning algorithm that dates back to the work of Fix and Hodges in 1951 \cite{hodges_fix}, and has grown to become one of the most popular machine learning techniques. The main idea is that a similarity of values in the feature space implies a similarity in the values for the target variable. Denote by $ x_{(1)},\dots,x_{(T)}$ a reordering of the time series entries, such that
\begin{equation} \label{eq:sorting}
    || x_{(1)} - x' ||  \le \dots \le || x_{(T)} - x' ||,
\end{equation} 
where $|| . ||$ denotes the Euclidean norm. The k-nearest neighbors set for the new entry $x$ is then defined as 
\begin{equation} \label{eq:nn} \textrm{NN}_k(x') = \{  i \in \{ 1,\dots,T  \} : || x_{(i)} - x' || \le || x_{(k)} - x' || \}. 
\end{equation}
Based on $\textrm{NN}_k(x')$, the forecast for $y'$ is computed as
\begin{equation} \label{eq:yhat_knn}  \hat{y} = \frac{1}{k} \sum_{i \in \textrm{NN}_k(x') } y_i. \end{equation} 
A closely related method is the Generalized Regression Neural Network (GRNN), introduced in \cite{specht}. When forecasting $y'$, GRNN attaches a weight to each of the previous time series entries, determined by their distance to $x'$. In particular, define the weight $w_i$ of the $i^{th}$ observation  as    
\begin{equation} \label{eq:w_i}
    w_i = \frac{\exp \left( \frac{-|| x_i - x' ||^2}{h}  \right)}{\sum\limits_{j=1}^{T} \exp \left( \frac{-|| x_j - x' ||^2}{h}  \right)},
\end{equation}

where $h$ is a bandwidth hyper-parameter to be optimized. The prediction for the response of a new observation $\hat{y}$ is then simply the weighted average of the responses of the training observations, defined as 
\begin{equation} \label{eq:yhat}
    \hat{y} = \sum_{i=1}^T w_i y_i.
\end{equation}
Consider the observations $(x_1,y_1),\dots,(x_T,y_T)$, to be realizations of an $\mathbb{R}^{d+1}$ random vector, and let $f$ denote the corresponding true underlying density of the random variable. 
The theoretical motivation of GRNN stems from the Parzen estimators \cite{specht}, used to estimate unknown probability density functions $f$ as
\begin{equation} \label{eq:prenzel}
    \hat{f}_h(x',y') = \frac{1}{T} \frac{1}{\left( \pi h \right)^{\frac{d+1}{2}}} \sum_{i=1}^T \textrm{exp} \left[ -\frac{(x'-x_i)^T (x'-x_i)}{h} \right] \textrm{exp} \left[-\frac{(y'-y_i)^2}{h} \right].
\end{equation}
If $h$ varies with $T$ such that $\lim_{T \to \infty} h(T) = 0$ and $\lim_{T \to \infty} Th^d(T) = 0$, these estimators are shown to be consistent, i.e
\begin{equation} \label{eq:consistent} plim_{T \to \infty} \hat{f}_{h(T)}(x',y') = f(x',y'), \ \ \forall (x',y') \in \mathbb{R}^{d+1}.  \end{equation}
Informally, \cite{specht} argues that if $\hat{f}(x,y) \approx f(x,y)$, then
\begin{equation} \label{eq:conditional}
\hat{y} = \sum_{i=1}^T w_i y_i = \int y' \frac{\hat{f}(x,y')}{ \int \hat{f}(x,y) dy} dy' \approx \int y' \frac{f(x,y')}{ \int f(x,y) dy} dy' =  \mathbb{E} \left[ y' | x' \right].
\end{equation}
We caution the reader that, despite GRNN bearing ``Neural Network" in its name, it is quite different from other popular variants. Specifically, aside from the width hyper-parameter, there are no ad-hoc hyperparameters to be chosen or optimized. Furthermore, it is a single-pass algorithm, as the weights are not optimized. Thus, unlike other neural network structures, it is much faster and easier to train but lacks the expressive power of neural networks. 

\subsection{Dimensionality reduction}
Both kNN and GRNN suffer from the so-called \emph{curse of dimensionality} which refers to the phenomenon that for certain methods, the amount of data points required to estimate an unknown function with a certain accuracy, grows exponentially with the dimensionality of the data set.\footnote{For a common illustration of this problem, consider points that lie uniformly on a $d$-dimensional hypercube of side length $(1+\epsilon)r$. In this setup, only a proportion of $(1+\epsilon)^{-d}$ data points will lie in the central inner hypercube of side length $r$, with the rest of the points lying in the outer shell. This demonstrates that the distance between data points becomes unsuitable for a meaningful description of the structure of high-dimensional data sets}
To alleviate this issue, kNN or GRNN is often applied after projecting the data points from $\mathbb{R}^d$ into a lower-dimensional space $\mathbb{R}^{p}$ ($p \le d$), while somewhat retaining the structure of the original data set. By far the most popular such method is Principal Component Analysis (PCA), which applies a linear transformation to the original features, chosen to maximize the variance of the projected features in the newly reduced space while ensuring the resulting features remain uncorrelated with each other.

Denote by $C(X)$ the covariance matrix of $X$, and its eigen-decomposition by $C(X) = U \Lambda U^T$, where $ \Lambda $ is a diagonal matrix capturing the $\lambda_1 \ge \dots \ge \lambda_d \ge 0$ eigenvalues of $C(X)$, and the columns of $U$ are given by the $u_1,\dots,u_d$ orthonormal eigenvectors (in the same order as their corresponding eigenvalues).
The principal component scores are given by the columns of $XU$, with the variance of each of the new components given by its corresponding eigenvalue (component $i$ has variance $\lambda_i$). Selecting the first $p$ components retains a proportion of the total variance in the feature matrix equal to $\sum_{i=1}^p \lambda_i / \sum_{i=1}^d \lambda_i$. Given the proportion of variance to retain $\delta \in (0,1)$, we set 
\begin{equation} \label{eq:p}
p = \textrm{min} \left\{ i: \frac{\sum\limits_{k=1}^i \lambda_k}{\sum\limits_{k=1}^d \lambda_k} \ge \delta \right\}.
\end{equation} 
We remark that random-matrix theory based techniques could also be employed for delimiting the informative part of the spectrum from the bulk of the eigenvalues that correspond to the noise component \cite{bouchaud_rmt}.
Note that before calculating $C(X)$, we ensure that $X$ is of full rank by removing any columns of $X$ that result in rank deficiency.\footnote{Using a QR-decomposition algorithm, we can express $X=QR$, where $Q$ is an orthogonal matrix and $R$ is an upper-triangular matrix. Then we remove any feature $i$, where $|R_{i,i}| < \epsilon$ for some small $\epsilon > 0$} Before applying PCA, we standardize $X$ (subtracting by each feature its mean and dividing by its standard deviation). We shall denote the transformed features as $\tilde{X} = XU$. 

\subsection{Maximal correlation}
Consider two univariate random variables denoted as $V_1$ and $V_2$. Loosely stated, the correlation between $V_1$ and $V_2$ is a measure of their association. Popular types of correlation include the Pearson, Spearman and Kendall correlations. Another type is the maximal correlation (see e.g  \cite{breiman}) defined as 
\begin{equation}\label{eq:max_cor}
    \underset{\phi,\psi \in \mathcal{F}}{\textrm{argmax}} \ \textrm{Cor}(\phi(V_1),\psi(V_2)),
\end{equation}
where $\mathcal{F}$ is a suitable function class and $\textrm{Cor}$ is the Pearson correlation defined as
\begin{equation} \label{eq:correlation}
    \textrm{Cor}(V_1,V_2) = \frac{\mathbb{E} \left[ V_1 V_2 \right] - \mathbb{E} \left[ V_1 \right] \mathbb{E} \left[ V_2 \right]}{\sqrt{\mathbb{E} \left[ V_1^2 \right] - \mathbb{E} \left[ V_1 \right]^2} \sqrt{\mathbb{E} \left[ V_2^2 \right] - \mathbb{E} \left[ V_2 \right]^2}}.
\end{equation} 
        
A solution to the problem of determining maximal correlation can be computed by optimizing the functions $\phi$ and $\psi$ iteratively. We consider a related but simpler problem, where we aim to calculate 
\begin{equation} \label{os_max_cor} 
    \underset{\phi \in \mathcal{F}}{\textrm{argmax}} \ \textrm{Cor}(\phi(V_1),V_2). 
\end{equation}
The reason for considering  \eqref{os_max_cor} instead of \eqref{eq:max_cor}, is that in a regression setting, we are only interested in transforming the feature, rather than the target. By solving \eqref{eq:max_cor}, one can transform the features in the embedding space so as to be maximally correlated to the target.

We consider a solution in a restricted function class $\mathcal{F}$.\footnote{If $\mathcal{F}$ is chosen to be the class of all functions, then the solution to the above problem can easily be shown to have the trivial solution $\Phi(V_1) = \frac{V_2-\mathbb{E} \left[ V_2 \right] }{\textrm{Var}(V_2)}$.} In particular, we consider the class of Bernstein polynomials (see e.g \cite{lorentz1986bernstein}), which are defined as 
\begin{equation}
    b_{\nu,n}(x) = \binom{\nu}{n} x^\nu (1-x)^{n-\nu}, \quad \nu = 0, 1, \dots, n. 
\end{equation} 
The choice of this basis is justified via the Stone-Weistrass theorem.\footnote{In its most usual form it says that for any real continuous function $f$ on the interval $\left[a,b \right]$ and $\epsilon > 0, \exists$ a polynomial function $g$ such that $\forall x \in \left[ a,b \right]$, we have $|f(x) - g(x)| < \epsilon$.} Bernstein polynomials have been widely used, especially in the context of Bezier curves \cite{Hosaka1992}, and B-splines \cite{Piegl1997}, often arising in 
the development of modeling tools in computer-aided geometric design.

For the solution of \Cref{os_max_cor}, we assume that we observe vectors $v_1$ and $v_2$, which capture $N$ observations of the random variables $V_1$ and $V_2$, respectively. Denote by $\bar{v}_1$ and $\bar{v}_2$ the mean of $v_1$ and $v_2$ respectively. Similarly denote by $\sigma_{v_1}$ and $\sigma_{v_2}$ standard deviation of $v_1$ and $v_2$ respectively.
Consider the function basis $\left( \phi_1, \dots, \phi_K \right)$, and define the $N \times K$ matrix $\Phi$ s.t $\Phi_{i,j} = \phi_j(v_{1,i})$. We assume $\Phi$ is centered, i.e $\Phi^T e_N = \textbf{0}$. 

With the function basis defined, we consider the function class 
\begin{equation} \label{F}
    \mathcal{F} = \set*{ \sum_{k=1}^{K}  c_k \phi_k : c_1,\dots,c_K \in \mathbb{R} },
\end{equation}  
so that for $f \in \mathcal{F}$, $f(v_{1,i}) = \sum_{k=1}^{K} c_k \phi_k(v_{1,i}),$
and problem \eqref{os_max_cor} reduces to finding the suitable vector $c$.

To simplify calculations, we also require the transformed vector to be standardized, i.e $ c^T \Phi^T e_N =0, \frac{1}{N-1} c^T \Phi^T \Phi c = 1$. Thus, we aim to solve 
\begin{equation} \label{eq:argmax}
     \underset{c \in \mathbb{R}^K,\ c^T \Phi^T e_N =0, \ c^T \Phi^T \Phi c = N-1}{\textrm{argmax}}\textrm{Cor}(\Phi c,v_2). 
\end{equation}

Note that \begin{equation} \label{eq:int1}\textrm{Cor}(\Phi c,v_2) = \frac{\textrm{Cov}(\Phi c,v_2)}{\sigma_{v_2} \sigma_{v_1}}  = \frac{\frac{(\Phi c)^T v_2}{N-1} - \left[ \frac{(\Phi c)^T e_N}{N} \right] \left[ \frac{v_2^T e_N}{N} \right]}{\sigma_{v_2} \sigma_{v_1}}, \end{equation}
and thus, under the constraints of \eqref{eq:argmax}, we have 
\begin{equation} 
\label{eq:int2} \textrm{Cor}(\Phi c,v_2) =  \frac{(\Phi c)^T v_2}{(N-1) \sigma_{v_2}}, 
\end{equation}
for which a solution can be found by considering the Lagrangian
\begin{equation} \label{lagrangian}
    L = \frac{ c^T \Phi^T v_2}{(N-1) \sigma_{v_2}} - \lambda (c^T \Phi^T e_N) - \mu (c^T \Phi^T  \Phi c - (N-1)). \end{equation}
Extrimizers of the above Lagrangian satisfy
\begin{equation} 
\label{eq:mu} \mu = \frac{ c^T \Phi^T v_2}{2(N-1)^2 \sigma_{v_2}}, \textrm{ and} 
\end{equation}
\begin{equation} \label{eq:lambda}  \lambda = \frac{\bar{v}_2}{(N-1) \sigma_{v_2}}. \end{equation}
Substituting back to \Cref{lagrangian}, and writing $z = \Phi^T v_2$, $\bar{z} = z - \bar{v}_2 \Phi^T e_N$ and $A = \frac{\Phi^ T \Phi}{N-1}$, we reduce the problem to solving \begin{equation}
    \bar{z} = \left( c^T z \right) A c. 
\end{equation}  
By considering a solution of the form $c = \tau A^{-1} \bar{z}$ for $\tau \in \mathbb{R}$, we arrive at the solution
\begin{equation} 
\label{eq:t} \tau = \frac{1}{\sqrt{\bar{z}^T (A^{-1})^T z}}. 
\end{equation}
Overall, the solution to \eqref{eq:argmax} is given by  
\begin{equation} \label{eq:c} c = \frac{A^{-1} \bar{z}}{\sqrt{\bar{z}^T (A^{-1})^T z}} = \frac{(\Phi^T \Phi)^{-1} \Phi^T \left( v_2 - \bar{v}_2 e_N \right)}{\sqrt{\left( v_2 - \bar{v}_2 e_N \right)^T \Phi \bar{z}^T \left[ (\Phi^T \Phi)^{-1} \right]^T \Phi^T v_2}}. \end{equation}
We will refer to the correlation between the target $y$ and the transformed $x$ as the \emph{one-sided maximal correlation} (OSMC) of x and $y$, as
\begin{equation} \label{eq:osmc}
    \textrm{OSMC}_K(x,y) = \textrm{Cor}(\Phi c,y),
\end{equation}   
where $\Phi$ and $c$ are depended on $x$ and the choice of the number of basis functions used $K$. 

\section{Methodology} \label{sec:main}
\begin{figure} 
   \includegraphics[width=1\textwidth]{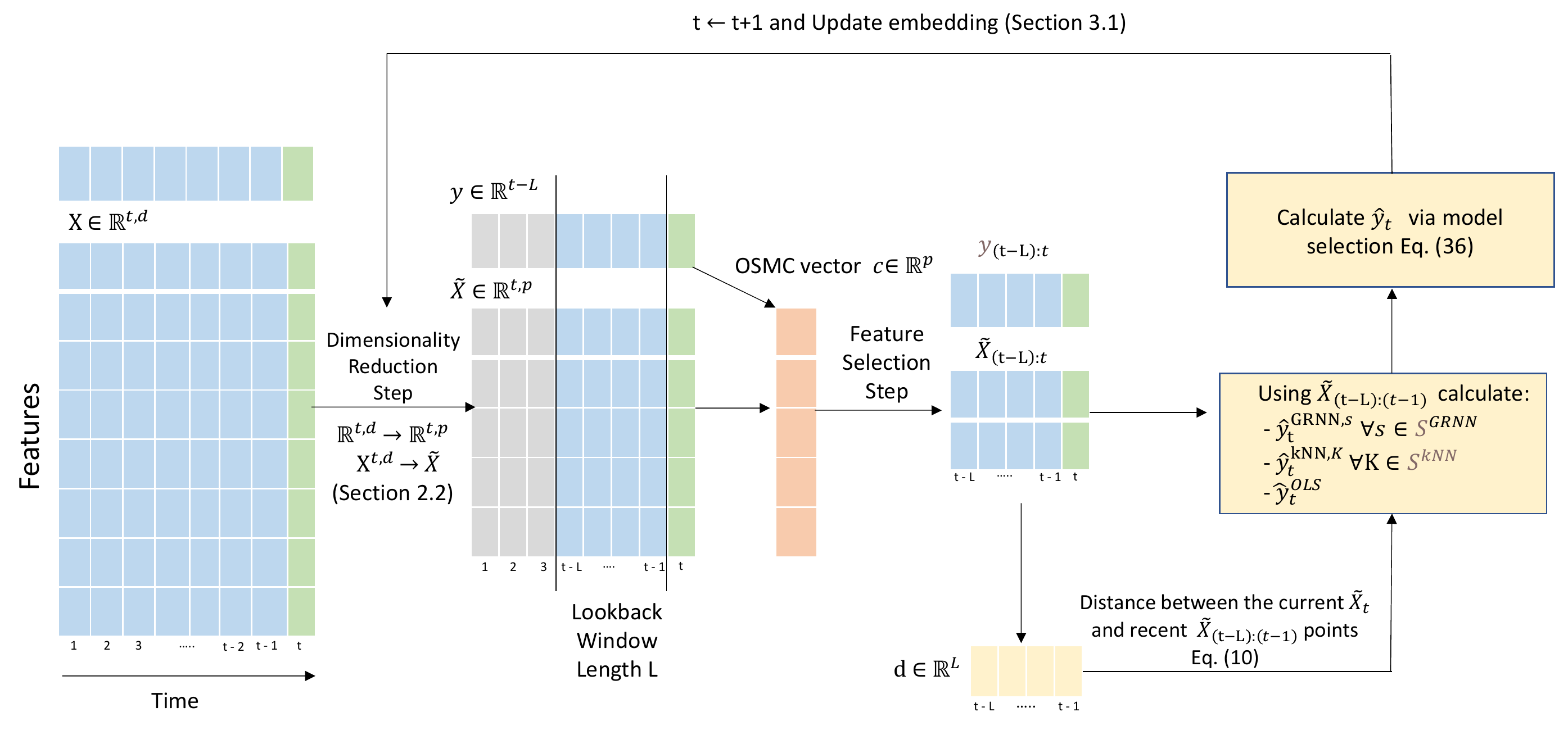}
   \caption{Pictorial illustration of {\ofter}, where the main components of its iterative part are displayed.}
   \label{fig:diagram}
\end{figure}

\Cref{fig:diagram} is a pictorial diagram of the main components of the {\ofter} pipeline, each of which is described in detail in the remainder of this section.

\subsection{Online update of the embedding}
\label{sec:update}
In order for the algorithm to run in an online setting, the embedding needs to be updated as new data arrives. The initial embedding will be calculated based on the first $L_0 < T$ entries, and then updated at every time step. However, re-computing the embedding at every single iteration can be quite expensive (in fact, of complexity $O(nd^2 + d^3)$ at each step). 
For PCA, there exists a computationally efficient approach that updates $U$ and $\Lambda$. In particular, if $C(X_t)$ denotes the covariance matrix built from the first $t$ data points, one can compute the eigenvectors of the updated covariance 
\begin{equation} \label{eq:cov_update}
    C(X_{1:t+1}) = \frac{t-1}{t} C(X_{1:t}) + \frac{1}{t} || x_{t+1} ||^2 \frac{x_{t+1}}{|| x_{t+1} ||} \frac{x_{t+1}^T}{|| x_{t+1} ||},
\end{equation} 
by using the previously computed eigenvectors of $C(X_{1:t})$ (we assume centered data for now). The online algorithm proposed in \cite{mitz2019roipca} achieves just that, with a complexity of $O(p^2 d)$ at each iteration (recall that $p$ denotes the chosen dimension of the embedding). 

The above is an instance of a more general problem, which is the rank-one update of an eigensystem \cite{secular}. Consider again the symmetric matrix $A$, whose eigen-decomposition is known, with $\lambda_1 \ge \dots \ge \lambda_d$ denoting its eigenvalues, and $u_i$ be the eigenvector corresponding to $\lambda_i, \forall i \in \{ 1,\dots,d \}$ such that $A = U \Lambda U^T$. 
The task at hand is to evaluate the eigensystem of $A + \rho v v^T$, by using the known eigensystem of $A$. As shown by \cite{secular}, the eigenvalues of $A + \rho v v^T$ are the roots $\kappa_1,\dots,\kappa_d$ of the \emph{secular equation}  
\begin{equation} \label{eq:secular} 1 + \rho \sum_{i=1}^d \frac{z_i^2}{\lambda_i - \kappa} = 0, \   \textrm{where}  \  z = U^T v. \end{equation}
Once the updated eigenvalues are computed, each updated eigenvector $u_k$ can be calculated as 
\begin{equation} 
u_k = \frac{U \left( \Lambda - \kappa_k I \right)^{-1} z}{||U \left( \Lambda - \kappa_k I \right)^{-1} z || }. \end{equation}
Going a step further, one could reduce computations by only updating the first few eigenvalues. A good approximation of the solution is given by \cite{mitz2019symmetric}, who shows how this can be achieved by solving 
\begin{equation} \label{eq:secular_m}
    1 + \rho \left( \sum_{i=1}^p \frac{z_i^2}{\lambda_i-\kappa} + \frac{1-\sum\limits_{i=1}^p z_i^2}{\mu-\kappa} \right) = 0. 
\end{equation}  
The solutions of this equation are shown to approximate the first $m$ eigenvalues, with an error of the order $O(\underset{m+1 \le i \le p}{\textrm{max}} |\lambda_i - \mu|)$. Note that $\mu$ is another parameter to be chosen, and should be made so as to minimize $O(\underset{m+1 \le i \le p}{\textrm{max}} |\lambda_i - \mu|)$, for example $\mu = \textrm{median}(\lambda_{m+1},\dots,\lambda_p)$. The parameter $p$ will have already been chosen at an earlier part of the algorithm using \Cref{eq:p}, and does not change during the update step.   
While there are various existing techniques to solve equation \Cref{eq:secular_m}, we employ the convexifying method described in \cite{melman}, adopting the earlier work of \cite{secular}.

Note that initially the feature matrix is standardized based on the initial $L_0$ entries (i.e we standardize $X_{(1:L_0)}$). Denote the mean vector of $X_{1:t}$ by $\mu_t \in \mathbb{R}^d$, and the correctly centered $X_{1:t}$ by $\bar{X}_{1:t} \in \mathbb{R}^d$ Assuming that $X_{(1:t-1)}$ has been centered by $\mu_{t-1}$, we need to re-center based on $\mu_{t}$. Following \cite{mitz2019roipca}, the update takes the form $\bar{X}_{1:t} = X_{1:t} + e_t (\mu_{t-1}-\mu_{t})^T$.
It turns out that we can adjust for the re-centering, via two more rank-one updates of the covariance matrix, by expressing $\bar{X}^T_{1:t} \bar{X}_{1:t}$ in the following way. 
\begin{align*}
    \bar{X}^T_{1:t} \bar{X}_{1:t} &= X_{1:t}^T X_{1:t} + X_{1:t}^T e_N (\mu_{t-1}-\mu_{t})^T + (\mu_{t-1}-\mu_{t}) e_N^T + t (\mu_{t-1}-\mu_{t}) (\mu_{t-1}-\mu_{t})^T \\
    &= X_{1:t}^T X_{1:t} + \begin{bmatrix} \mu_{t-1}-\mu_{t}  & X_{1:t}^T e_N \end{bmatrix}  \begin{bmatrix} t & 1 \\ 1 & 0 \end{bmatrix}  \begin{bmatrix} (\mu_{t-1}-\mu_{t})^T  \\ e_N^T X_{1:t} \end{bmatrix} \\
    &= X_{1:t}^T X_{1:t} + \begin{bmatrix} \mu_{t-1}-\mu_{t}  & X_{1:t}^T e_N \end{bmatrix} \begin{bmatrix} \frac{\rho_1}{\sqrt{1 + \rho_1^2}} & \frac{\rho_2}{\sqrt{1 + \rho_2^2}} \\ \frac{1}{\sqrt{1 + \rho_1^2}} & \frac{1}{\sqrt{1 + \rho_2^2}} \end{bmatrix} \begin{bmatrix} \rho_1 & 0 \\ 0 & \rho_2 \end{bmatrix} \begin{bmatrix} \frac{\rho_1}{\sqrt{1 + \rho_1^2}} & \frac{\rho_2}{\sqrt{1 + \rho_2^2}} \\ \frac{1}{\sqrt{1 + \rho_1^2}} & \frac{1}{\sqrt{1 + \rho_2^2}} \end{bmatrix}^T\begin{bmatrix} (\mu_{t-1}-\mu_{t})^T  \\ e_N^T X_{1:t} \end{bmatrix}\\
        &= X_{1:t}^T X_{1:t} + \begin{bmatrix} \frac{\rho_1 (\mu_{t-1}-\mu_{t}) + X_{1:t}^T e_N}{\sqrt{1 + \rho_1^2}}  & \frac{\rho_2 (\mu_{t-1}-\mu_{t}) + X_{1:t}^T e_N} {\sqrt{1 + \rho_2^2}}\end{bmatrix}  \begin{bmatrix} \rho_1 & 0 \\ 0 & \rho_2 \end{bmatrix} \begin{bmatrix} \frac{\rho_1 (\mu_{t-1}-\mu_{t}) + X_{1:t}^T e_N}{\sqrt{1 + \rho_1^2}}  \\ \frac{\rho_2 (\mu_{t-1}-\mu_{t}) + X_{1:t}^T e_N} {\sqrt{1 + \rho_2^2}} \end{bmatrix} \\
        &= X_{1:t}^T X_{1:t} + \rho_1 b_1 b_1^T + \rho_2 b_2 b_2^T,
  \end{align*}
where in the above 
\begin{equation} \label{eq:rho1}
    \rho_1 =\frac{t - \sqrt{t^2 + 4}}{2},
\end{equation}
\begin{equation} \label{eq:rho2}
    \rho_2 = \frac{t + \sqrt{t^2 + 4}}{2},
\end{equation}
\begin{equation} \label{eq:b}
    b_1 = \frac{\rho_1 (\mu_{t-1}-\mu_{t}) + X_{1:t}^T e_N}{\sqrt{1 + \rho_1^2}},
\end{equation}
\begin{equation} \label{eq:cc}
    b_2=\frac{\rho_2 (\mu_{t-1}-\mu_{t}) + X_{1:t}^T e_N} {\sqrt{1 + \rho_2^2}}.
\end{equation}

\subsection{Choice of scaling factor}
The hyperparameter $k$ of kNN in \eqref{eq:yhat} and the hyperparameter $h$ in \eqref{eq:w_i} are both of crucial to
the performance of their respective algorithms. Notice that, as $h \to \infty$,  $w_i \to 1/T$ (\Cref{eq:w_i}) for all $i$. On the other hand, as $h \to 0$, GRNN becomes equivalent to  $1\textrm{NN}$ (kNN with $k=1$). To demonstrate the effect of the scaling factor on the predictive task, we consider the following toy model
\begin{equation} \label{eq:toy}
    y_t = 0.5 + \cos \left( \frac{\pi}{64} t \right) + \epsilon_t, \textrm{where  } \epsilon_t \sim N(0,\sigma^2).
\end{equation}
\begin{figure}[h]
   \includegraphics[width=1\textwidth]{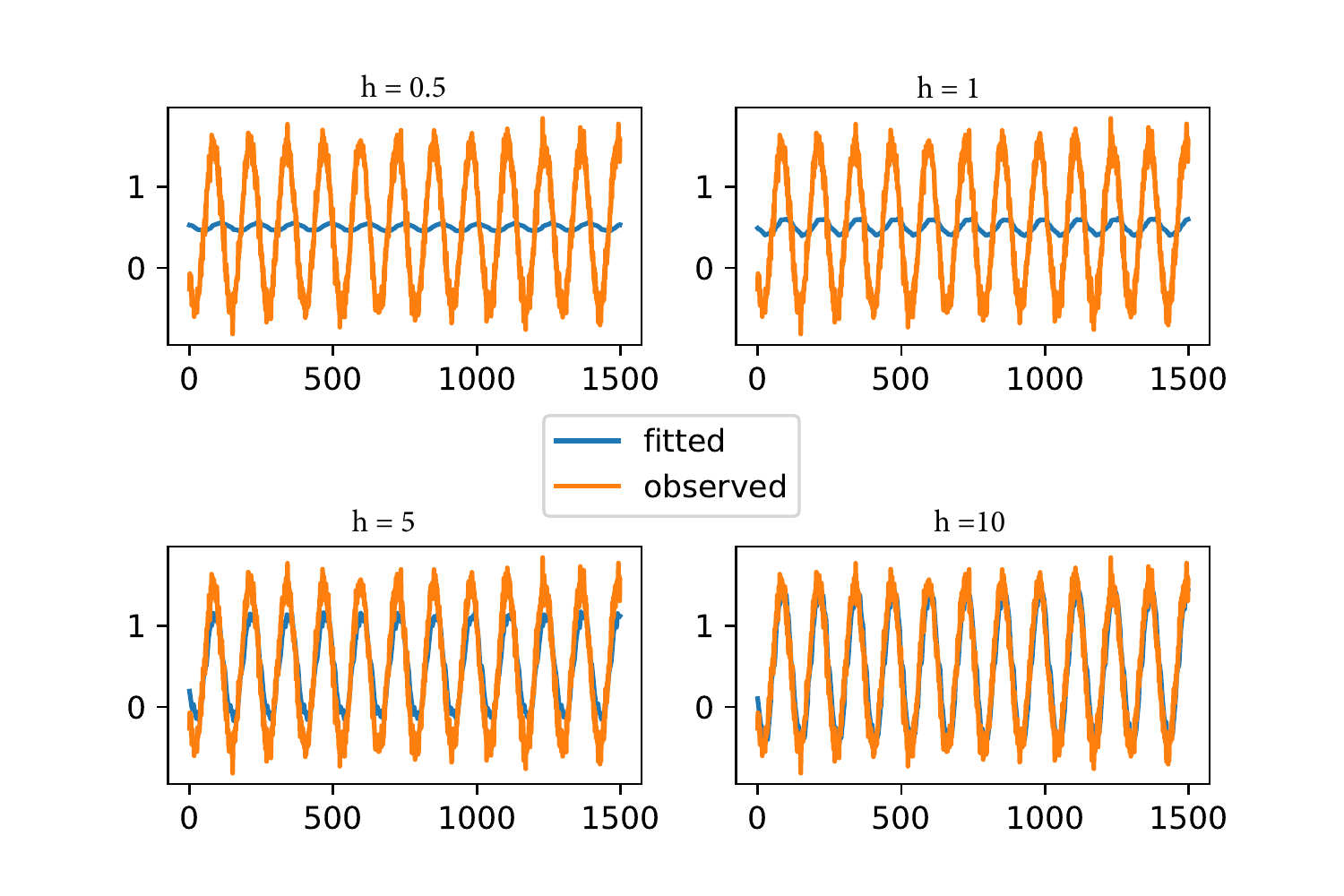}
   \caption{Visualization of the forecasted values when using GRNN with different scaling parameters for the model \eqref{eq:toy}. In a high signal-to-noise problem like this, as $h$ becomes larger GRNN becomes almost perfectly capable of forecasting future values.  
   }
   \label{fig:toy_model}
\end{figure}
In \Cref{fig:toy_model} we demonstrate how the choice of the scaling factor affects the nature of the obtained fit. For low values of $h$, the obtained fit is close to giving constant fitted values equal to the mean, while for high scaling factors the obtained fit matches the observed data points almost perfectly.\footnote{Note that this is a pattern for this specific toy data set and not a general pattern}

To choose a suitable value of $h$, \cite{specht} proposes the use of the holdout method, where the optimal parameter is found by calculating a loss function by removing a sample point at a time and taking the average. More modern applications use K-fold validation instead, and these methods can be modified for time series data (see \cite{BERGMEIR} for a discussion on the different approaches). 

To remove any scale dependencies, we re-parameterize $h$ as \begin{equation} \label{eq:h }h = \frac{\textrm{median}(\{ || x_j - x' ||  \}_{j=1}^T)}{s}. \end{equation} 
Since this is an online algorithm, we could update the scaling parameter accordingly. Instead of choosing a single bandwidth parameter from a candidate set, we use each of the candidate bandwidths to fit a different model, updating the performance of all models at each time point. The model with the highest performance up to the previous time point is chosen for the forecast of the current time point.\footnote{Alternatively, we can take a weighted average of the forecasts of each model, but the performance was inferior.}

The choice of the scaling factor can be performed by the use of a loss function. If the fitted value corresponding to scaling parameter $s$ for day $t$ is $\hat{y}^s_t$, the forecast $\hat{y_t}$ is calculated as
\begin{equation} \label{eq:yhat_comb} \hat{y}_t = \sum_{s=1}^{|S|} \eta_t^s \hat{y}_t^s, \ \textrm{where} \ \  
\eta_t^s = 
\begin{cases}
      1, \ \  &  R_t^s = \textrm{min}(R_t^1,\dots,R_t^S) \\
      0, \ \  &  \textrm{otherwise}\\
\end{cases}.
\end{equation}
By $R_t^s$, we define the loss corresponding to parameter $s$, i.e $R_t^s = \sum_{i=1}^{t-1} \textrm{Loss}(\hat{y}_t^s,y_t)$. The loss function can be the Mean Squared Error (MSE) or Mean Absfolute Error (MAE) for general applications. For the application of forecasting future returns, we use the loss function of minus P\&L (we define P\&L in \Cref{sec:finance}).

Similarly, a large of $k$ for kNN algorithm would produce similar forecasts to using low values of $h$. Instead of using either GRNN or kNN, we use both models, each with their own hyperparameter candidate set, and then select based on performance, in a similar manner to \eqref{eq:yhat_comb}. Aside from these two types of models, we also consider a standard OLS model. 

Denote by $S^{\textrm{GRNN}}$ the hyperparameter canditate for GRNN and $S^{\textrm{kNN}}$ the hyperparameter canditate set for kNN. Denote by $\hat{y}^{\textrm{GRNN},s}_t$, the forecast when using GRNN with parameter $s$, $\hat{y}^{\textrm{kNN},k}_t$ the forecast when using kNN (with parameter $k$), and $\hat{y}_t^{\textrm{OLS}}$ the forecast when using an OLS model.
Then define
\begin{equation} \label{eq:R_init}
  \begin{gathered}
   R^{\textrm{GRNN},s}_t = \sum_{i=1}^{t-1} \textrm{Loss}(\hat{y}^{\textrm{GRNN},s}_i,y_i). \\
   R^{\textrm{kNN},k}_t = \sum_{i=1}^{t-1} \textrm{Loss}(\hat{y}^{\textrm{kNN},k}_i,y_i). \\
      R^{\textrm{OLS}}_t = \sum_{i=1}^{t-1} \textrm{Loss}(\hat{y}^{\textrm{OLS}}_i,y_i).
  \end{gathered}
\end{equation}
Based on the losses up to time point $t$, the minimum loss is calculated as
\begin{equation}
R^{\textrm{optimal}}_t = \textrm{min}(\{ R^{\textrm{GRNN},s}_t: s \in S^{\textrm{GRNN}} \} \cup \{ R^{\textrm{kNN},k}_t: k \in S^{\textrm{kNN}} \} \cup \{ R_t^{\textrm{OLS}} \}),
\end{equation}
and the number of models achieving the optimal performance as
\begin{equation}
    N^{\textrm{optimal}}_t = |\{ s: R^{\textrm{GRNN},sx} = R^{\textrm{optimal}}_t \} |  + |\{ k: R^{\textrm{kNN},k} = R^{\textrm{optimal}}_t\} | + \mathbbm{1}({R_t^{\textrm{OLS}} = R^{\textrm{optimal}}_t}),
\end{equation}
where $\mathbbm{1}(A)$ is an indicator function taking the value $1$, if condition $A$ is satisfied and $0$ otherwise.  
Then we reformulate \eqref{eq:yhat_comb}, to include the kNN and OLS models as
\begin{equation} \label{eq:yhat_comb_new} \hat{y}_t = \sum_{s \in S^{\textrm{GRNN}}} \eta_t^{\textrm{GRNN},s} \hat{y}^{\textrm{GRNN},s}_t
+ \sum_{k \in S^{\textrm{kNN}}} \eta_t^{\textrm{kNN},k} \hat{y}_t^{\textrm{kNN},k} + \eta_t^{\textrm{OLS}}  \hat{y}_t^{\textrm{OLS}}, 
\end{equation}
where 
\begin{equation} \label{eq:eta_init}
  \begin{gathered}
  \eta^{\textrm{GRNN},s}_t = 
\begin{cases}
      \frac{1}{ N^{\textrm{optimal}}_t}, \ \  &  R^{\textrm{GRNN},s}_t =R^{\textrm{optimal}}_t \\
      0, \ \  &  \textrm{otherwise}\\
\end{cases}, \\
\eta^{\textrm{kNN},k}_t = 
\begin{cases}
      \frac{1}{ N^{\textrm{optimal}}_t}, \ \  &  R^{\textrm{kNN},k}_t = R^{\textrm{optimal}}_t \\
      0, \ \  &  \textrm{otherwise}\\
\end{cases}, \\
    \eta^{\textrm{OLS}}_t = 
\begin{cases}
      \frac{1}{ N^{\textrm{optimal}}_t}, \ \  &  R^{\textrm{OLS}}_t =  R^{\textrm{optimal}}_t  \\
      0, \ \  &  \textrm{otherwise}\\
\end{cases}.
  \end{gathered}
\end{equation}
    
\subsection{Overcoming the shortcomings of kNN and GRNN}
One downside of applying dimensionality reduction methods is that they are agnostic of the target to be forecasted. In transforming the initial space in order to retain a large proportion of the variability with a reduced number of features, an initial feature that is already highly correlated with the target but also weakly correlated with other features could end up having a relatively small weight in the final forecasting step. To circumvent this shortcoming, we employ two additional steps.

Firstly, we make sure that all features from the original feature space $X$ that have a high enough correlation with the target are retained in the forecasting step. This way, we ensure that selecting the components that capture most of the variability is not done at the expense of dropping original features that provide predictive power on their own.
Therefore, we define the following set, after calculating the OSMC between each of the original features and the target 
\begin{equation} \label{eq:augment}
I^{\textrm{Original}} = \{ j:  \textrm{OSMC}(X_{1:L_0, j },y_{1:L_0}) \ge c_{\textrm{Original}} \}.
\end{equation}
During the pre-processing part of the algorithm, we replace $\tilde{X}$ with $(\tilde{X},X_{1:T,I^{\textrm{Original}}})$, with a number of features equal to $p + |I^{\textrm{Original}}|$. Note that for mid-sized, low signal-to-error data sets like the financial data set we use in \Cref{sec:finance}, $I_{\textrm{Original}}$ is usually null. 

Secondly, certain features in the embedding space ought to have more influence on the neighbor selection. This can be achieved by using a weighted metric rather than the standard Euclidean norm. In particular, if $v$ represents the vector of weights, the weighted Euclidean distance between $X_{t_1}$ and $X_{t_2}$ is given by  
\begin{equation} \label{eq:distance} || \tilde{X}_{t_1} - \tilde{X}_{t_2} ||_v = \sqrt{\sum_{j=1}^{p + |I^{\textrm{Original}}|} v_j (\tilde{X}_{t_1,j} - \tilde{X}_{t_2,j})^2}. 
\end{equation}
The weights are based on the OSMC between each feature and the target
\begin{equation} \label{eq:v}
     v_j = \frac{c_j^2 \mathbbm{1}(|c_j | \ge c_{\textrm{min}})}{\sum\limits_{l=1}^{p + |I^{\textrm{Original}}|}  c_l^2 \mathbbm{1}(| c_l | \ge c_{\textrm{min}})},
\end{equation}
where $c_j = \textrm{OSMC}(\tilde{X}_{1:L_0, j },y_{1:L_0}) \ \  \forall j \in \{ 1, \dots, p, p + 1, \dots, |I^\textrm{Original}| \}$.

\subsection{Algorithm outline} \label{sec:outline}
To ensure stationarity of the target $y$, we first perform an ADF test with significance level $p_{\textrm{ADF}}$, and if the null hypothesis is not rejected, we difference the time series, and repeat until we reject the null hypothesis. We then perform ACF and PACF tests to check if there are any significant ARMA effects.\footnote{We use a significance level of 0.001. We consider a specific lag effect significant if the confidence interval of either the ACF or the PACF is above or below zero.} If such effects are identified, we perform a grid search over the orders of the AR and MA orders (p and q respectively) to find the optimal ARIMA(p,r,q) model, where $r$ denotes the number of times differencing was applied. Otherwise, we simply fit ARIMA(0,r,0).

Using this procedure, we decompose the target $y$ into a time series and a residual component
\begin{equation} \label{eq:y_time_series}
    y = \underbrace{y^{\textrm{TS}}}_{\textrm{Time series component}} + \underbrace{\epsilon}_{\textrm{Residual}}.
\end{equation} 
Overall, the algorithm consists of three main steps: (1) computing an embedding, (2) choosing features based on the OSMC, and (3) forecasting based on kNN and GRNN. The complete algorithm is given in \Cref{alg:a2}.

\begin{algorithm} 
\KwIn{$\tilde{X}$ (Feature Matrix), $y$ (Target), $L$ (Lookback Window Length), $v$ (distance vector), $S^{\textrm{kNN}}, S^{\textrm{GRNN}}$ (parameter candidate sets)} 
\KwOut{$\hat{y}_t$} 
\caption{Forecasting for time point $t$} \label{alg:a1}

\begin{algorithmic}[1]
\STATE Compute $d_{j,t} = || \tilde{X}_{ t, I} - \tilde{X}_{ j, I }||_v$ for $j=t-L,\dots,t-1$ 
\FOR{$s \in S^{\textrm{GRNN}}$}
\STATE Set $h_s = \frac{\textrm{median}(\{ d_{j,t} \}_{j=t-L}^{t-1})}{s}$
\STATE Compute $w_{j,t}^s = \frac{\exp \left(\frac{- d_{j,t}^2}{h_s}  \right)}{\sum\limits_{m=t-L}^{t-1} \exp \left(\frac{- d_{m,t}^2}{h_s}  \right)} \forall j \in \{ t-L,\dots,t-1\}$
\STATE Compute $\hat{y}^{\textrm{GRNN},s} = \sum\limits_{j=t-L}^{t-1} w_{j,t}^s y_i$
\ENDFOR
\FOR{$k \in S^{\textrm{kNN}}$}
\STATE Find the set $\textrm{NN}_k(\tilde{X}_{t,})$ of the $k$-nearest neighbors of $\tilde{X}_{t}$, in $\tilde{X}_{(t-L):(t-1)}$
\STATE Compute $\hat{y}^{\textrm{kNN},k} = \frac{1}{k} \sum\limits_{i \in \textrm{NN}_k(\tilde{X}_{t}) } y_i$ 
\ENDFOR 
\STATE Compute $\hat{y}^{\textrm{OLS}} = \beta_0 + \beta^T \tilde{X}_{t}$
\STATE Estimate the fitted value $\hat{y}_t$ using \eqref{eq:yhat_comb_new}
\end{algorithmic}
\end{algorithm}

\begin{algorithm} 
\KwIn{$X$ (Feature Matrix), y (target), $L_0$ (Training Window Length), $L$ (Lookback Window Length), $c_{\textrm{min}}$ , $c_{\textrm{Original}}$ (correlation thresholds), $S^{\textrm{kNN}}, S^{\textrm{GRNN}}$ (parameter candidate sets), $\delta$ (proportion of variance to retain),$p_{\textrm{ADF}}$ (p-value for ADF), LOSS (loss function)} 
\KwOut{$\hat{y}$} 
\caption{OFTER} \label{alg:a2}
\begin{algorithmic}[1]

\STATE Use the procedure described in \Cref{sec:outline}, to arrive at the optimal ARIMA($p$,$r$,$q$) model for $y$ 
\STATE Use the optimal model to obtain fitted values $y^{\textrm{TS}}$. If ARIMA(0,0,0) is chosen set $y^{\textrm{TS}} = (0,\dots,0)$
\STATE Transform $X$ so as to make $X_{1:L_0}$ a full column-rank (by QR decomposition) and standardized matrix 
\STATE Obtain the first $p$ principal components of $X_{ 1:L_0 }$, with $p$ chosen via \Cref{eq:p}, and 
set $U_p = (u_1,\dots,u_p)$ and $\Lambda_p = (\lambda_1,\dots,\lambda_p)$
\STATE Obtain the $L_0 \times p$ matrix $\tilde{X} = XU_{1:L_0,1:p}$
\STATE Set $I^{\textrm{Original}} = \{ i:  \textrm{OSMC}(X_{1:L_0, i},y_{1:L_0}) \ge c_{\textrm{Original}} \}$
\STATE Replace $\tilde{X}$ by $(\tilde{X}, X_{1:T,I^{\textrm{Original}}})$
                    \STATE Compute $R^{\textrm{GRNN},s}_{L_0} \ \forall s \in S^{\textrm{GRNN}}, R^{\textrm{kNN},k}_{L_0} \ \forall k \in S^{\textrm{kNN}}$, and $R^{\textrm{OLS}}_{L_0}$ using  \eqref{eq:R_init} \\
\STATE Compute $\eta^{\textrm{GRNN},s}_{L_0} \ \forall s \in S^{\textrm{GRNN}}, \eta^{\textrm{kNN},k}_{L_0} \ \forall k \in S^{\textrm{kNN}}$, and $\eta^{\textrm{OLS}}_{L_0}$ using  \eqref{eq:eta_init} \\
    \FOR{$j \gets 1$ to $p + |I^{\textrm{Original}}|$}
   \STATE Compute $c_j = \textrm{OSMC}(\tilde{X}_{1:L_0, j },y_{1:L_0})$
   \STATE  Set $v_{j} = \frac{c_j^2 \mathbbm{1}(|c_j | \ge c_{\textrm{min}})}{\sum_{l=1}^{p + |I^{\textrm{Original}}|}  c_l^2 \mathbbm{1}(| c_l | \ge c_{\textrm{min}})}$
\ENDFOR
\STATE Regress $y_{1:L_0}$ on $\tilde{X}_{1:L_0,}$ using OLS regression, to obtain the coefficient vector $\beta$ and the intercept $\beta_0$
\FOR{$t \gets L_0+1$ to $D$}
\STATE Compute $\hat{y}_t$ using   \Cref{alg:a1}
\STATE Update $U_p$ and $\Lambda_p$, using  the secular equation method (detailed in \Cref{sec:update})
\STATE If t is divisible by 100, re-fit the OLS model using $\tilde{X}_{(t-L):t}$ and $y_{(t-L):t}$, to obtain $\beta$ and $\beta_0$ 
\ENDFOR 
\STATE Update $\hat{y}$ to $\hat{y} + y^{\textrm{TS}}_{(L_0+1):D}$
\end{algorithmic}
\end{algorithm}

\section{Empirical results} \label{sec:results}
We consider and compare four different configurations of {\ofter}, which we denote as follows
\begin{itemize}
\item {\ofter} (no dimensionality reduction and no feature transformation). This entails setting $\tilde{X} = X$, using the Pearson correlation instead of the OSMC, and skipping steps 4,5, 6, 7 and 17 of \Cref{alg:a2}. 
\item {\ofter}-DR (dimensionality reduction and no feature transformation). This is almost the same as \Cref{alg:a2}, but we use the Pearson correlation instead of the OSMC. 
\item {\ofter}-FT  (feature transformation and no dimensionality reduction). This entails setting $\tilde{X} = X$ as well as skipping steps 4,5, 6, 7 and 17 of \Cref{alg:a2}. 
\item {\ofter}-DR-FT (feature transformation and dimensionality reduction). This is the exact algorithm as given by \Cref{alg:a2}. 
\end{itemize}

Unless stated otherwise we use $p_{\textrm{ADF}} = 0.05$. When performing dimensionality reduction, we consider $\delta = 0.9$. For the candidate sets we use $S^{GRNN} = \{ 0.001,0.005,0.01,0.05,0.1,0.5,1,5,10,50,100 \}$ and 
$S^{kNN} = \{1,2,3,5,10,15,20,30,50\}$. In terms of features, we include those corresponding to the current period up to a lag of $3$. Lastly, we set $L_0$ equal to $ \floor*{ 0.7 T}$. Note that {\ofter} is generally robust for different choices of hyper-parameters, as evidenced by our robustness analysis reported in \Cref{sec:robust}.

\subsection{Baseline models}
We compare the performance of {\ofter} against that of several mainstream machine learning models. We include the linear models OLS, PCR, LASSO, but also the widely used SVM and Random Forests. We also include some of the most widely used time series tailored deep learning models, such as LSTM, Bi-LSTM, and GRU. Unlike {\ofter}, most of these methods are not naturally amenable to efficient implementation in an online setting and thus cannot be updated. For comparison purposes, we retrain each of these models every 100 steps. In the case of LSTM, Bi-LSTM, and GRU, we re-fit the model with only five epochs, starting from the previous weight matrix. For the sake of a fair comparison, instead of following the embedding updating step, we also re-compute the embedding of {\ofter} every 100 steps. We provide an overview of these methods, as well as details of their implementation for this task, in \Cref{sec:appendix}. 

\subsection{Synthetic data sets}
Before testing {\ofter} on real life data sets, we first check whether it can perform consistently well on some simple low dimensional toy models. For this purpose, the three five-dimensional models used by \cite{papaioannou2021time}, which are simplistic,  that are crafted to resemble electroencephalography signals, which is suitable as a data set for this forecasting task. 

\begin{equation}  \label{eq:m1_synthetic}
\textrm{M1:}
\begin{cases}
 y_t^{(1)} = 0.2 y_{t-1}^{(1)} - 0.4 y_{t-1}^{(2)} + \epsilon_{t}^{(1)} \\
 y_t^{(2)} = -0.5 y_{t-1}^{(1)} + 0.15 y_{t-1}^{(2)} + \epsilon_{t}^{(2)} \\ 
 y_t^{(3)} = -0.14 y_{t-1}^{(2)} + \epsilon_{t}^{(3)} \\ 
 y_t^{(4)} = 0.5 y_{t-1}^{(1)} - 0.25 y_{t-1}^{(2)} + \epsilon_{t}^{(4)} \\
 y_t^{(5)} = 0.15 y_{t-1}^{(1)} + \epsilon_{t}^{(5)} 
\end{cases}
\end{equation}
The first data set (M1) shown in \Cref{eq:m1_synthetic} is generated by a set of five linear equations.

\begin{equation} \label{eq:m2_synthetic}
\textrm{M2:}
\begin{cases}
 y_t^{(1)} = 3.4 y_{t-1}^{(1)} \left(1-\textrm{exp} \left( -\left( y_{t-1}^{(1)} \right)^2 \right) \right) \textrm{exp}\left({-y_{t-1}^{(1)}}^2 \right)  + \epsilon_{t}^{(1)} \\ 
 y_t^{(2)} = 3.4 y_{t-1}^{(2)} \left(1-\textrm{exp} \left( - \left( y_{t-1}^{(2)} \right)^2 \right) \right) \textrm{exp}\left({-y_{t-1}^{(2)}}^2 \right) + 0.5 y_{t-1}^{(1)} y_{t-1}^{(2)} + \epsilon_{t}^{(2)} \\
 y_t^{(3)} = 3.4 y_{t-1}^{(3)} \left(1-\textrm{exp} \left( - \left(y_{t-1}^{(3)}\right)^2  \right) \right) \textrm{exp}\left({-y_{t-1}^{(3)}}^2 \right) + 0.3 y_{t-1}^{(2)} + 0.5 {y_{t-1}^{(1)}}^2 + \epsilon_{t}^{(3)} \\
 y_t^{(4)} = 0.5 y_{t-1}^{(1)} - 0.25 y_{t-1}^{(2)} + \epsilon_{t}^{(4)}  \\
 y_t^{(5)} = 0.15 y_{t-1}^{(1)} + \epsilon_{t}^{(5)} \end{cases}
\end{equation}
The second data set (M2) shown in \Cref{eq:m2_synthetic} is generated by a set of five equations, some of which are non-linear. \cite{papaioannou2021time} based these equations on some models appearing in \cite{constantinidou}. 

\begin{equation}  \label{eq:m3_synthetic}
\textrm{M3:}
\begin{cases}
y_t^{(1)} = 0.1 y_{t-1}^{(1)} - 0.6 y_{t-3}^{(2)}  + \epsilon_{t}^{(1)} \\
y_t^{(2)} = -0.15 y_{t-3}^{(1)} + 0.8 y_{t-3}^{(2)} + \epsilon_{t}^{(2)} \\
y_t^{(3)} = -0.45 y_{t-3}^{(2)} + \epsilon_{t}^{(3)} \\ 
y_t^{(4)} = 0.45 y_{t-3}^{(1)} - 0.85 y_{t-3}^{(2)} + \epsilon_{t}^{(4)} \\
y_t^{(5)} = 0.95 y_{t-2}^{(1)} + \epsilon_{t}^{(5)} 
\end{cases}
\end{equation}
The third data set (M3) resembles M1, but is a time series with a higher order, making M3 a VAR model.

In each of the above cases $\epsilon_t^{(1)},\dots,\epsilon_t^{(5)}$ are uncorrelated normally distributed random variables with unit variance. $y_1^{(1)},\dots,y_1^{(5)}$ are also equal to zero.

\subsection{Real data sets} \label{sec:finance}

\subsubsection*{Financial equity data set}
We consider a financial data set for two forecasting applications, namely daily returns and volumes. We consider a universe of 210 instruments, which consists of 10 ETFs (SPY, XLF, XLB, XLK, XLV, XLI, XLU, XLY, XLP, XLE), and 200 equities from the financial sector. 

Let $P_{i,d}$ denote the closing price for asset $i$ at day $d$. We consider the close to close returns\footnote{The returns considered are adjusted for splits and dividends} for asset $i$, defined as 
\begin{equation} \label{eq:return}
    R_{i,d}^{k} = \frac{P_{i,d+k}-P_{i,d}}{P_{i,d}}.
\end{equation}   

As a more representative quantity, grounded within financial applications, we consider the market excess returns, that is the difference between the returns of an instrument and that of the S\&P-500 index, a widely used proxy for market returns.   
\begin{equation} \label{eq:return_excess} \tilde{R}_{i,t}^{k} = \tilde{R}_{i,t}^{k} - \tilde{R}_{\textrm{SPY},t}^{k}.  
\end{equation}  

Denote the volume of instrument $i$ as $V_{i,t}$, which is defined as the total number of shares traded during day $t$. Note that this quantity is non-stationary and generally increases over time, across most instruments. 
To this end, we instead consider the 'log-volume-returns', defined as
\begin{equation} \label{eq:volume} 
\delta^{k}_{i,t} = \frac{\textrm{log}(V_{i,t+k})-\textrm{log}(V_{i,t})}{\textrm{log}(V_{i,t})}. 
\end{equation} 
Predicting daily returns is one of the most popular yet difficult problems in finance. 
Due to the low signal-to-noise ratio of this task, standard metrics like Mean Squared Error and correlation, are considered unsuitable. A more suitable economic-driven evaluation method is following a hypothetical transaction in the underlying, based on the forecasted signs (with $+$ and $-$ corresponding to buy and sell respectively). As this is simply for evaluation purposes, we will ignore transaction costs and liquidity constraints. 

Following this hypothetical strategy, we consider the (excess-returns-adjusted) Profit and Loss (P\&L) metric defined by
\begin{equation} \label{eq:PnL}
\textrm{P}\&\textrm{L}^{k}_t = \frac{1}{N} \sum_{i=1}^{N_t} \widetilde{R}^{k}_{i,t} \times \textrm{sign} \left(  s_{i,t} \right),
\end{equation} 
\vspace{0mm}
The P\&L is defined above for each day $d$, which yields a vector  
$\textbf{P\&L} = \left( \textrm{P\&L}_1,\textrm{P\&L}_2,\dots,\textrm{P\&L}_{T-1}\right)$ of daily P\&L's, which measures the outcome of the hypothetical strategy over the entire time horizon under consideration. A measure of the long-term performance of a strategy is the annualized Sharpe Ratio (hereafter SR) \cite{Shapiro1994}, defined as
\begin{equation}
\textrm{SR} = \frac{\textrm{mean}(\textbf{P\&L}) \times \sqrt{252}}{\textrm{sd}(\textbf{P\&L})}.
\end{equation}   
Lastly, we define the P\&L per dollar traded (PPD) as 
\begin{equation}
    \textrm{PPD} = \frac{1}{T-1} \sum_{t=1}^{T-1} \textrm{P\&L},
\end{equation} 
which assumes that every day, the total allocated capital to the portfolio remains constant at 1 USD.

We employ the SR significance test by \cite{bailey2014}. As returns are known to exhibit heavier tails than a normal distribution, this test is an extension of a traditional $t$-test, adjusted to account for the skewness and kurtosis of the distribution of the returns.

A higher magnitude forecast should naturally signify a higher confidence. For each trading day, we consider the quantiles of the forecast magnitudes $( \{|s_{i,t}| \}_{i=1}^N$, in order to define five quintile portfolios, of decreasing size. In particular, we define Q1 to be the portfolio consisting of all assets, Q2 to be the portfolio consisting of the top 80\% assets with the highest magnitude, and so on, up to Q5 which consists of the 20\% highest magnitude assets.

The feature matrix consists of the following indicators
\begin{itemize}
    \item 1, 2 and 3-day returns for each of the instruments, i.e $R^1_{i,t}$, $R^2_{i,t}$, $R^3_{i,t}$ 
    \item 1, 2 and 3-day absolute returns for each of the instruments, i.e $|R^1_{i,t}|$, $|R^2_{i,t}|$, $|R^3_{i,t}|$ 
    \item 1, 2 and 3-day log-volume returns for each of the instruments, i.e $\delta^1_{i,t}$, $\delta^2_{i,t}$,  $\delta^3_{i,t}$
    \item the three Fama \& French daily market factors
    \item the daily returns of 49 different industry portfolios\footnote{The industry portfolio returns and the daily market factors are taken directly from the Kenneth R. French library \cite{fama_french}. Based on a market portfolio, the three factors include small minus big (SMB) based on firm size, high minus low (HML) based on the book-to-market ratio, and the market excess return of the portfolio.}
    \item daily returns of the VIX index
\end{itemize}
We build an embedding based on all the above features. We set the signal $s_{i,t}$ equal to the output of the {\ofter} algorithm for each of the instruments $i$ and day $t$. 

\subsubsection*{Air quality data set}
As a non-financial application we also consider the Beijing air quality data set from the UCI Machine Learning Repository. Data is collected by the Beijing Municipal Environmental Monitoring Center, and considers various air quality measurements from 12 monitoring points. 
We aim at forecasting each of the six types of pollution (PM2.5,PM10,03,C0,S02,N02) concentrations (in $ug/m^3$), for each of the stations. Other features in the data set, used as predictors, include Temperature, Dewpoint Temperature, Precipitation, Pressure as well as Wind speed and Direction.

\subsection{Numerical results}
\begin{table}[h]
\resizebox{\textwidth}{!}{%
\centering
\begin{tabular}{l|llllllllllllll}
\hline
\textbf{Data Set} & \textbf{T} & \textbf{d} & \textbf{LM} & \textbf{PCR} & \textbf{LASSO} & \textbf{\begin{tabular}[c]{@{}l@{}}Random\\  Forests\end{tabular}} & \textbf{SVM} & \textbf{LSTM} & \textbf{Bi-LSTM} & \textbf{GRU} & \textbf{OFTER} & \textbf{OFTER-DR} & \textbf{OFTER-FT} & \textbf{OFTER-DR-FT} \\ \hline
Synthetic         & 3000       & 15         & 0.01        & 0,02         & 0.64           & 33.90                   & 1.81         & 173.45        & 202.99           & 184.81       & 7.87           & 9.68              & 37.52             & 24.53                \\ \hline
Financial         & 5022       & 154        & 0.37        & 0.48         & 5.38           & 3,144.02                & 0.17         & 442.86        & 503.78           & 549.30       & 17.58          & 22.25             & 222.87            & 153.06               \\ \hline
Air Quality       & 1461       & 228        & 0.10        & 0.31         & 2.23           & 132.05                  & 1.00         & 165.48        & 152.22           & 184.85       & 3.25           & 20.18             & 157.95            & 62.64                \\ \hline
\end{tabular}} \\ \\ 
 \caption{The average running time of all 12 methods on all of the forecasted quantities for each of the three multivariate time series. $T$ denotes the length of the time series, while $d$ denotes the dimensionality of each data set.}
   \label{tab:times}
\end{table}
We start with a discussion  of the running time of {\ofter}. In \Cref{tab:times}, we provide the average running time of the algorithm for each of the three considered data sets. We notice that all four configurations of {\ofter} are significantly faster than LSTM, Bi-LSTM and GRU, while at the same time significantly slower than the OLS, LASSO, PCR and SVM, as expected.
It is also clear that among the {\ofter} configurations, {\ofter}-FT is the slowest as each feature is transformed separately without first reducing the dimensionality of the feature space. The time complexity of {\ofter} and {\ofter}-DR are more comparable, with the latter being slightly more computationally intensive. 

For the numerical experiments concerning forecasting returns, we employ economic-driven metrics typical of this literature, as discussed in \Cref{sec:finance}. For all the remaining experiments, we base the performance comparisons on the (Pearson) correlation metric, as it is easily interpretable, in comparison to mean squared error and other similar metrics.\footnote{Note that the comparison between the performances of different models is almost identical if we instead use the mean squared error or the mean absolute error.}

\subsubsection*{\bb \; Synthetic data} 
For this task we use $L = 800, C_\textrm{min} = 0.05$ and $C_\textrm{Original} = 0.05$. In this experiment, {\ofter} performs consistently well. In almost all of the features, one of the {\ofter} configurations is among the two best-performing algorithms (with the GRU generally having a very high performance). 

\begin{figure}[h]
   \includegraphics[width=1\textwidth]{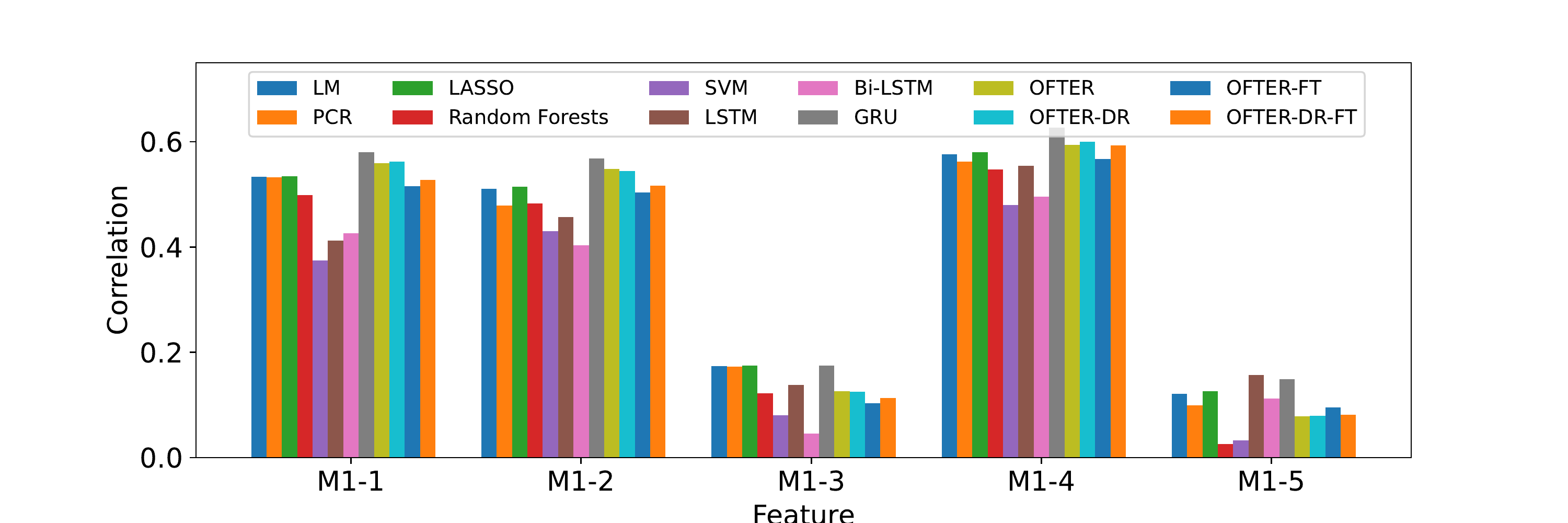}
   \caption{Grouped barplot showing a performance comparison of the 12 methods considered, on each of the five features of the M1 data set described in \eqref{eq:m1_synthetic}. The metric of the comparison is the correlation between the target values and the forecasts. 
   }
   \label{fig:synthetic1}
\end{figure}

The results on the synthetic data set confirm the expectations from our model. On M1, GRU appears to outperform the other methods for four out of five features as shown in \Cref{fig:synthetic1}. Following GRU, {\ofter} and {\ofter}-DR appear to have the highest performance for three out of five features. 

\begin{figure}[h]
   \includegraphics[width=1\textwidth]{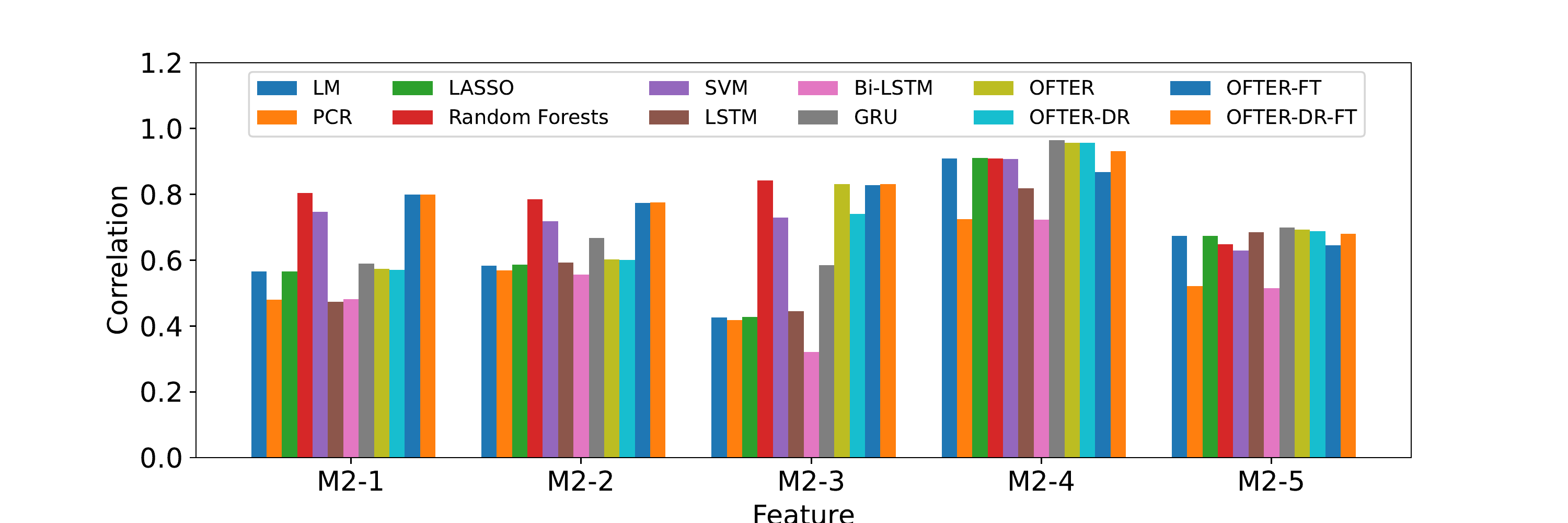}
   \caption{Grouped barplot showing a performance comparison of the 12 methods considered, on each of the five features of the M2 data set described in \eqref{eq:m2_synthetic}. The metric of the comparison is the correlation between the target values and the forecasts. 
   }
   \label{fig:synthetic2}
\end{figure}

In the case of M2, which is a non-linear multivariate non-linear model, the configurations of {\ofter} that employ the feature transformation component, are outperforming all other models, with the exception of Random Forests, in three out of five features (in particular, the three features which are defined in a non-linear manner) as can be seen in \Cref{fig:synthetic2}.
\begin{figure}[h]
   \includegraphics[width=1\textwidth]{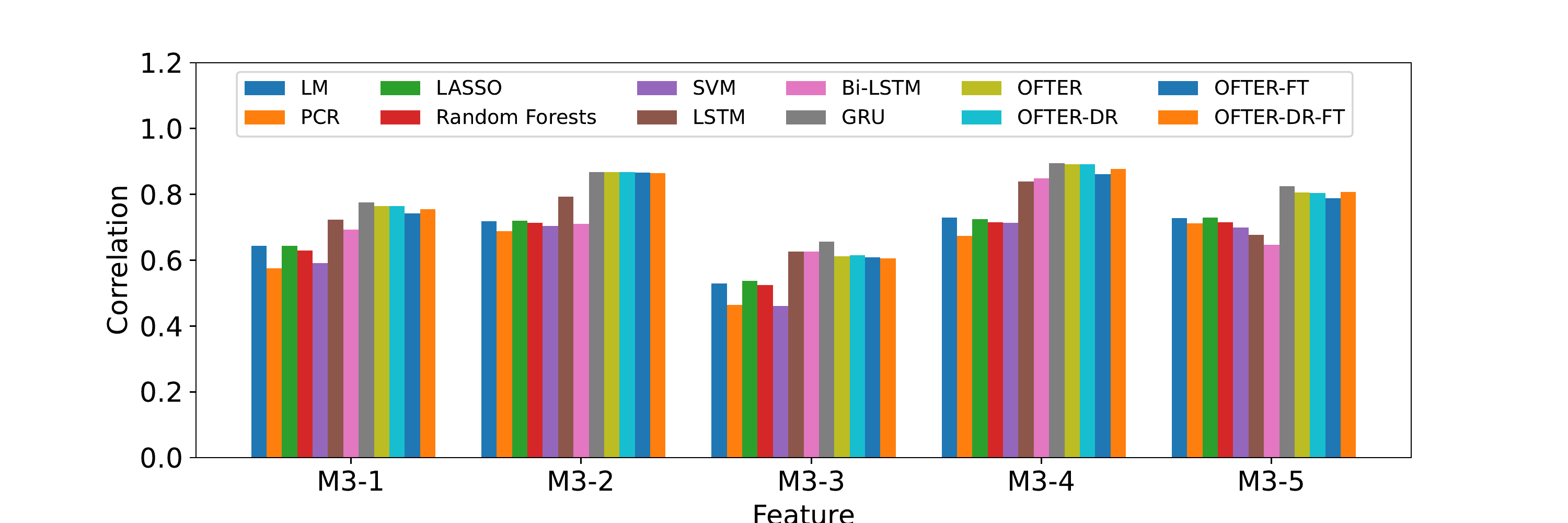}
   \caption{Grouped barplot showing a performance comparison of the 12 methods considered, on each of the five features of the M3 model described in \eqref{eq:m3_synthetic}. The metric of the comparison is the correlation between the target values and the forecasts.}
   \label{fig:synthetic3}
\end{figure}
Lastly, in the case of M3, which regards a vector autoregressive (VAR) model, {\ofter} is performing particularly well, outperforming all other models, with the exception of GRU, as can be seen in \Cref{fig:synthetic3}. Similarly to M1, the two {\ofter} configurations that do not employ the feature transformation component, fare slightly better.

\subsubsection*{\bb \; ETF volumes}
For both financial tasks, we use $L=800$ and set $C_\textrm{min} = C_\textrm{Original} = 0.01$, as typically, these are problems with a lower signal-to-noise ratio. 
\begin{figure}[h]
\includegraphics[width=1\textwidth]{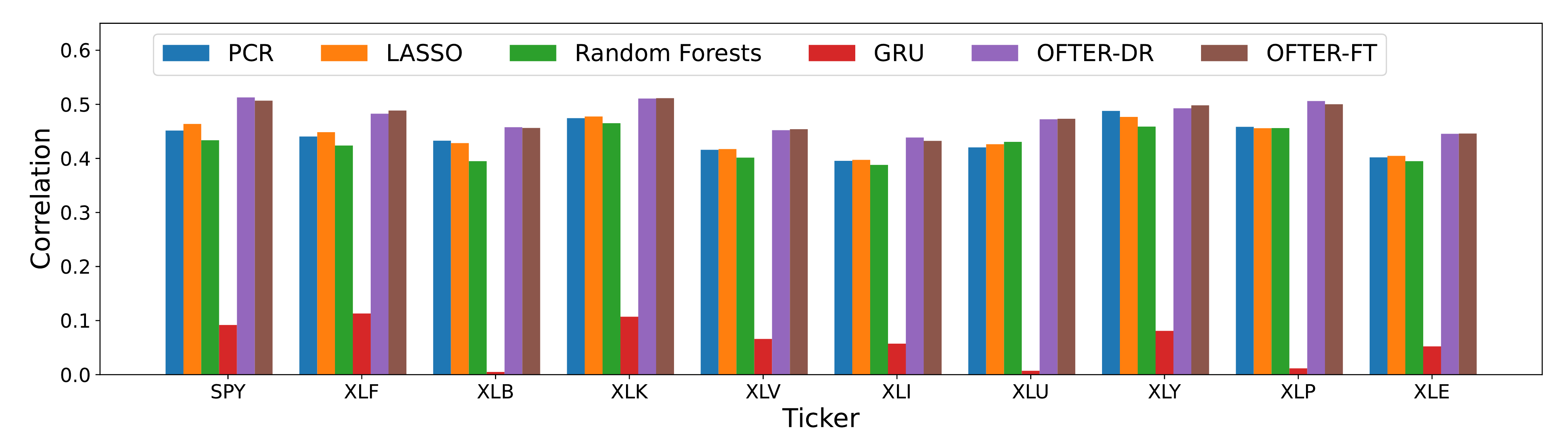}
   \caption{Groupped barplot showing the correlation between the delta-volume and its forecasts, by a selection of the baselines and {\ofter} configurations. The selection was such that the highest performing methods were selected from each category of methods.
   }
   \label{fig:barvolumes}
\end{figure} 
For the task of predicting the log-delta-volumes of ETF instruments \eqref{eq:volume}, {\ofter} outperforms all the other benchmark methods. As demonstrated in \Cref{fig:barvolumes}, both  {\ofter}-FT and {\ofter}-DR configurations outperform all of the baseline methods, across all 10 ETFs.
\subsubsection*{\bb \; ETF market excess returns} 
\begin{table}[]
\resizebox{\textwidth}{!}{%
\centering
\begin{tabular}{@{}c|ccc|ccc|ccc|ccc|ccc@{}}
                         & \multicolumn{3}{c|}{Q1}                                                                    & \multicolumn{3}{c|}{Q2}                                                                    & \multicolumn{3}{c|}{Q3}                                                                    & \multicolumn{3}{c|}{Q4}                                                                    & \multicolumn{3}{c}{Q5}                                       \\
\cellcolor[HTML]{FFFFFF} & \cellcolor[HTML]{FFFFFF}SR & \cellcolor[HTML]{FFFFFF}PPD & \cellcolor[HTML]{FFFFFF}p-value & \cellcolor[HTML]{FFFFFF}SR & \cellcolor[HTML]{FFFFFF}PPD & \cellcolor[HTML]{FFFFFF}p-value & \cellcolor[HTML]{FFFFFF}SR & \cellcolor[HTML]{FFFFFF}PPD & \cellcolor[HTML]{FFFFFF}p-value & \cellcolor[HTML]{FFFFFF}SR & \cellcolor[HTML]{FFFFFF}PPD & \cellcolor[HTML]{FFFFFF}p-value & \cellcolor[HTML]{FFFFFF}SR & PPD            & p-value        \\ \midrule
LM                       & -0.745                     & -1.211                      & 0.967                           & -0.729                     & -1.270                      & 0.964                           & -0.873                     & -1.783                      & 0.984                           & -0.740                     & -1.689                      & 0.965                           & -0.814                     & -2.664         & 0.977          \\
PCR                      & -0.863                     & -1.324                      & 0.981                           & -0.901                     & -1.523                      & 0.985                           & -0.841                     & -1.737                      & 0.979                           & -0.952                     & -2.234                      & 0.989                           & -0.809                     & -2.728         & 0.976          \\
LASSO                    & -0.890                     & -1.293                      & 0.985                           & -0.920                     & -1.411                      & 0.987                           & -1.052                     & -1.905                      & 0.995                           & -1.192                     & -2.466                      & 0.998                           & -0.995                     & -3.163         & 0.993          \\
Random Forests                       & 0.371                      & 0.534                       & 0.182                           & 0.282                      & 0.436                       & 0.246                           & 0.174                      & 0.339                       & 0.335                           & 0.233                      & 0.521                       & 0.285                           & 0.285                      & 1.000          & 0.243          \\
SVM                      & -0.370                     & -0.528                      & 0.807                           & -0.314                     & -0.471                      & 0.763                           & 0.309                      & 0.429                       & 0.204                           & 0.613                      & 0.984                       & 0.058                           & \textbf{0.706}             & \textbf{1.267} & \textbf{0.039} \\
LSTM                     & 0.308                      & 0.433                       & 0.227                           & 0.261                      & 0.396                       & 0.262                           & 0.421                      & 0.771                       & 0.152                           & 0.403                      & 0.818                       & 0.163                           & 0.059                      & 0.176          & 0.442          \\
Bi-LSTM                  & -0.810                     & -0.868                      & 0.976                           & -0.824                     & -1.017                      & 0.977                           & -0.388                     & -0.641                      & 0.828                           & -0.238                     & -0.462                      & 0.719                           & -0.456                     & -1.377         & 0.866          \\
GRU                      & 0.424                      & 0.610                       & 0.152                           & 0.286                      & 0.448                       & 0.243                           & 0.148                      & 0.268                       & 0.359                           & 0.144                      & 0.294                       & 0.363                           & 0.129                      & 0.403          & 0.376          \\
OFTER                    & 0.534                      & 0.811                       & 0.089                           & 0.653                      & 1.082                       & 0.052                           & 0.616                      & 1.203                       & 0.064                           & 0.290                      & 0.646                       & 0.233                           & 0.444                      & 1.463          & 0.140          \\
OFTER-DR                 & \textbf{1.616}             & \textbf{2.346}              & \textbf{0.000}                  & \textbf{1.704}             & \textbf{2.607}              & \textbf{0.000}                  & \textbf{1.229}             & \textbf{2.194}              & \textbf{0.001}                  & \textbf{1.258}             & \textbf{2.441}              & \textbf{0.001}                  & \textbf{1.261}             & \textbf{3.765} & \textbf{0.001} \\
OFTER-FT                 & \textbf{1.143}             & \textbf{1.737}              & \textbf{0.002}                  & \textbf{1.344}             & \textbf{2.197}              & \textbf{0.000}                  & \textbf{1.261}             & \textbf{2.476}              & \textbf{0.001}                  & \textbf{0.988}             & \textbf{2.167}              & \textbf{0.007}                  & 0.491                      & 1.545          & 0.110          \\
OFTER-DR-FT              & \textbf{1.468}             & \textbf{2.125}              & \textbf{0.000}                  & \textbf{1.532}             & \textbf{2.409}              & \textbf{0.000}                  & \textbf{1.562}             & \textbf{2.795}              & \textbf{0.000}                  & \textbf{0.884}             & \textbf{1.760}              & \textbf{0.015}                  & \textbf{0.826}             & \textbf{2.371} & \textbf{0.022}
    \end{tabular}} \\
\caption{In this table we summarize the performance of 12 models for the ETF price movement forecasting task. SR denotes the annualized Sharpe Ratio, PPD denotes the Profit Per Dollar, and p-value denotes the p-value obtained from the one-tailed SR significance test of \cite{bailey2014}. Q1-Q5 correspond to the quantile portfolios as defined in \Cref{sec:finance}.}  \label{tab:etf_results}
\end{table}

\begin{figure}[h]
\centering
\includegraphics[trim={0cm 0cm 0cm 1.5cm},width=1\textwidth]{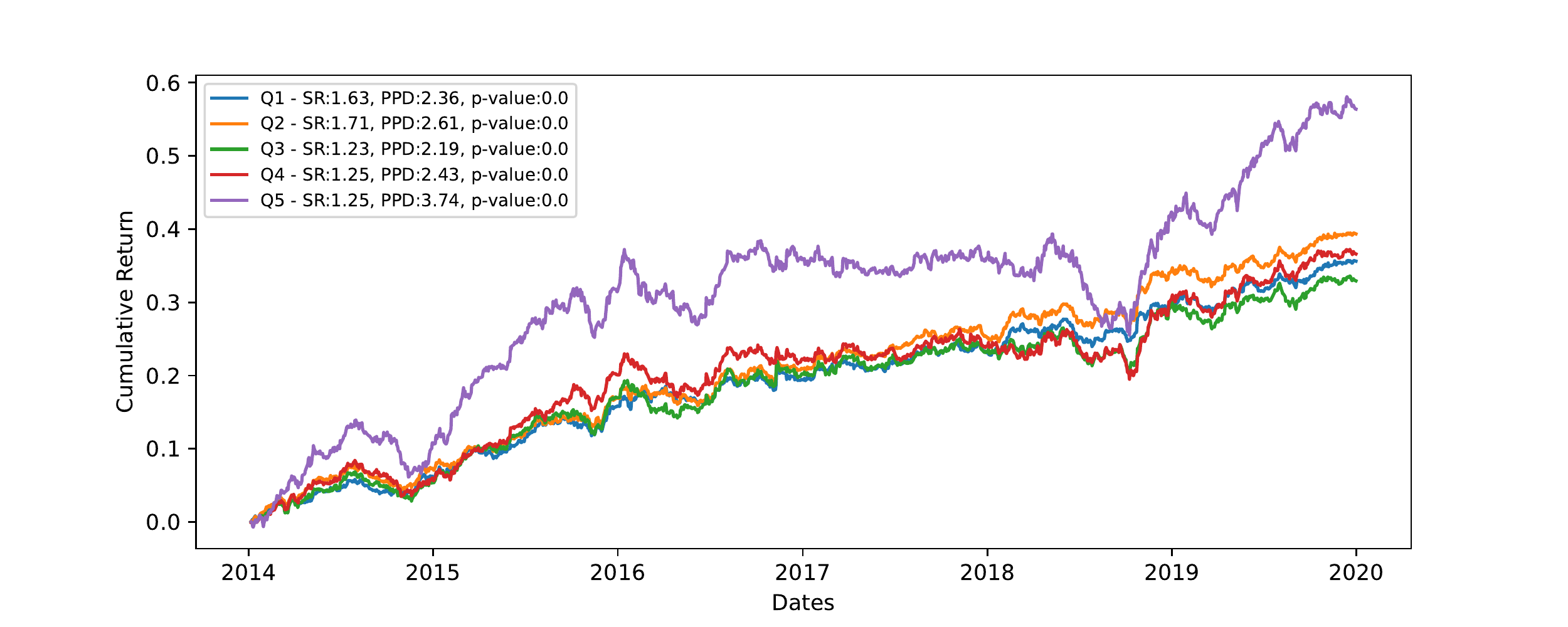}
   \caption{Cumulative P\&L plot for the daily OFTER-DR generated strategy. SR denotes the annualized Sharpe Ratio, PPD denotes the Profit Per Dollar, and p-value denotes the p-value obtained from the one-tailed SR significance test of \cite{bailey2014}. Q1-Q5 correspond to the quantile portfolios as defined in \Cref{sec:finance}.}   
   \label{fig:cums_ofter}
\end{figure}

\begin{figure}[h]
\centering
\includegraphics[trim={0cm 0cm 0cm 0cm},width=1\textwidth]{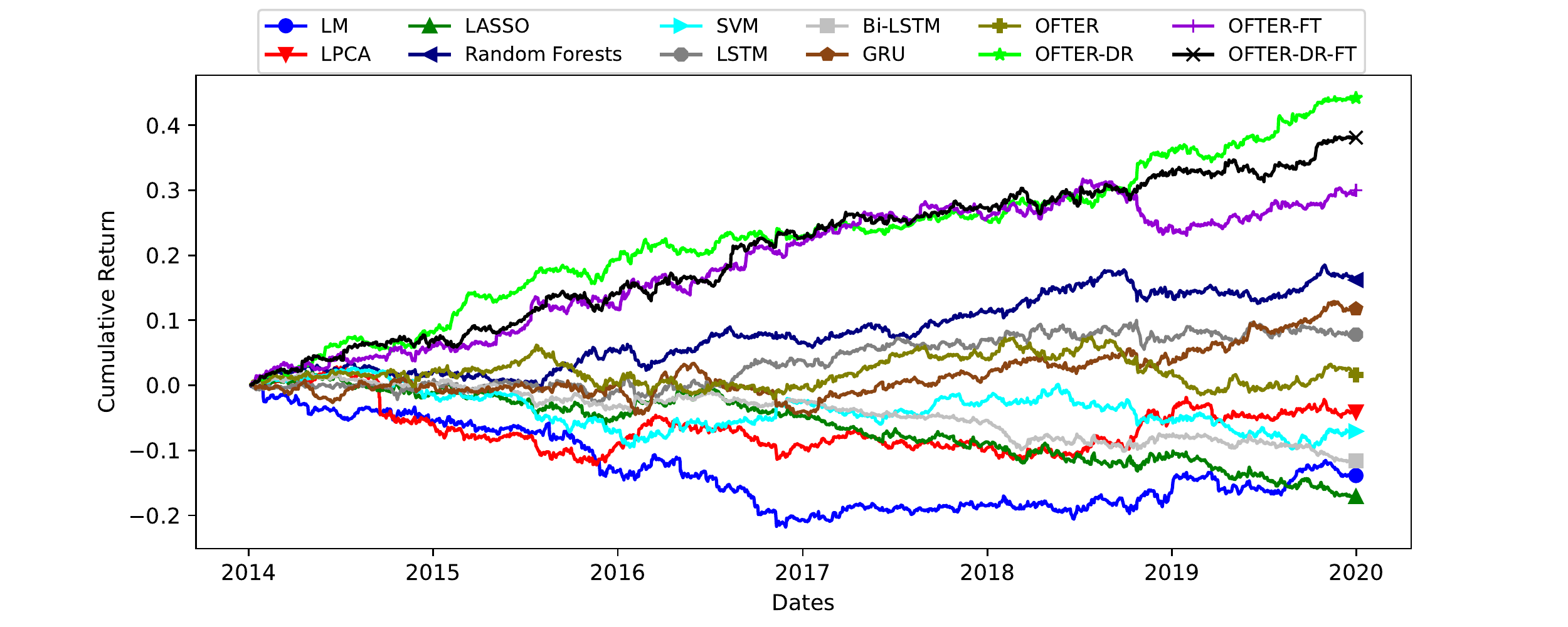}
\caption{Cumulative P\&L plot of the daily strategies corresponding to the  12 methods for the ETF price movement forecasting task. SR denotes the annualized Sharpe Ratio, PPD denotes the Profit Per Dollar, and p-value denotes the p-value as obtained from the one-tailed Sharpe Ratio significance test of \cite{bailey2014}. The performance metrics for each method are given in \Cref{tab:etf_results}.}  
   \label{fig:cums_methods}
\end{figure} 

We simulate a daily strategy based on trading ten ETF instruments for the Q1-Q5 portfolios,  and  summarize the results in \Cref{tab:etf_results}. For three out of four {\ofter} strategies, the SR is considered statistically significant.\footnote{Using the Sharpe Ratio significance test of \cite{bailey2014}. This is based on a generalization of the t-test, that accounts for the non-normality of returns, by taking into account the kurtosis and skewness of the P\&L distribution.} As a visual demonstration of the results, we also plot the cumulative P\&L for the Q1 strategy of each method in \Cref{fig:cums_methods}, clearly demonstrating that OFTER-DR and OFTER-DR-FT are outperforming the other methods. With the exception of the Q5-SVM strategy, none of the other benchmark methods result in strategies with a statistically significant SR. Across most quantiles, OFTER-DR scores the highest performance, both in terms of SR and PPD. In \Cref{fig:cums_ofter}, we plot the cumulative P\&L for the OFTER-DR strategy, clearly demonstrating a consistently well-performing set of strategies across different quantiles.

\subsubsection*{\bb \; Air pollution data set}
\begin{figure}[ht!]
   \includegraphics[width=1
\textwidth]{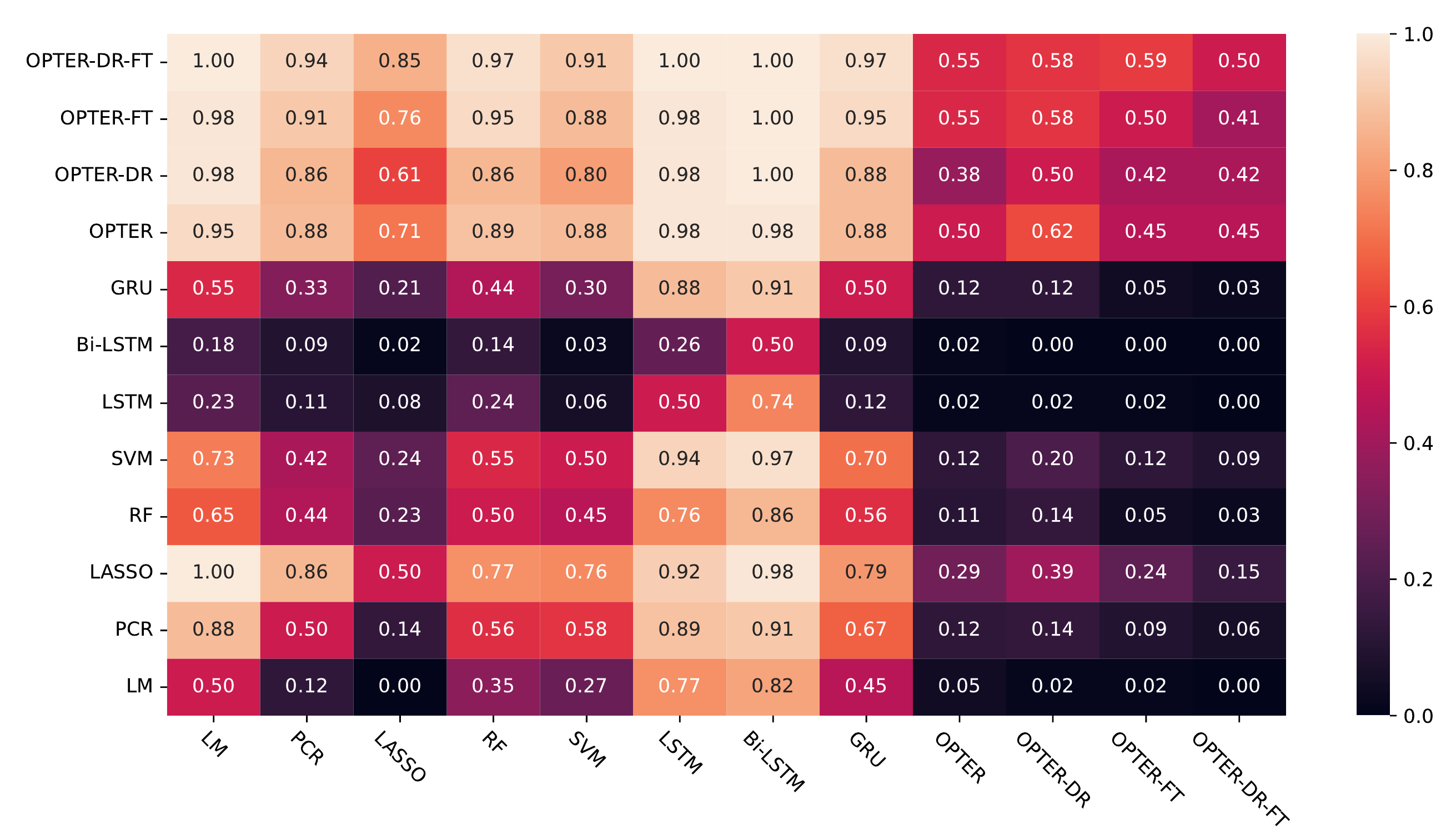}
   \caption{This heatmap is a visualization of the comparison between the different forecasting models for the air pollution data set. The proportion of the number of cases where a  method (label of the vertical axis) outperforms another method (label of the horizontal axis) in terms of correlation is given in the corresponding cell. 
   }
   \label{fig:pollution}
\end{figure} 

Finally, for the air pollution task, we forecast a total of 6 features, for 12 stations. In \Cref{fig:pollution}, we show a grid of the proportions of cases where one method outperforms another based on the correlation. As it can be seen, all four {\ofter} configurations are outperforming all of the other eight benchmarks. In this particular case, {\ofter}-DR-FT outperforms the other three versions of {\ofter}.

\section{Applications to feature importance and outlier detection.} 
\label{sec:benefits}
In this section, we discuss  additional applications of {\ofter} to two tasks of interest in the machine learning community. In particular, we consider a feature importance component that boosts the interpretative power of {\ofter}, and an application to outlier detection.

\subsection{Feature importance}
\label{sec:feature_importance}
\begin{figure}[h]
   \includegraphics[trim={5cm 4cm 6cm 4.5cm},width=1\textwidth]{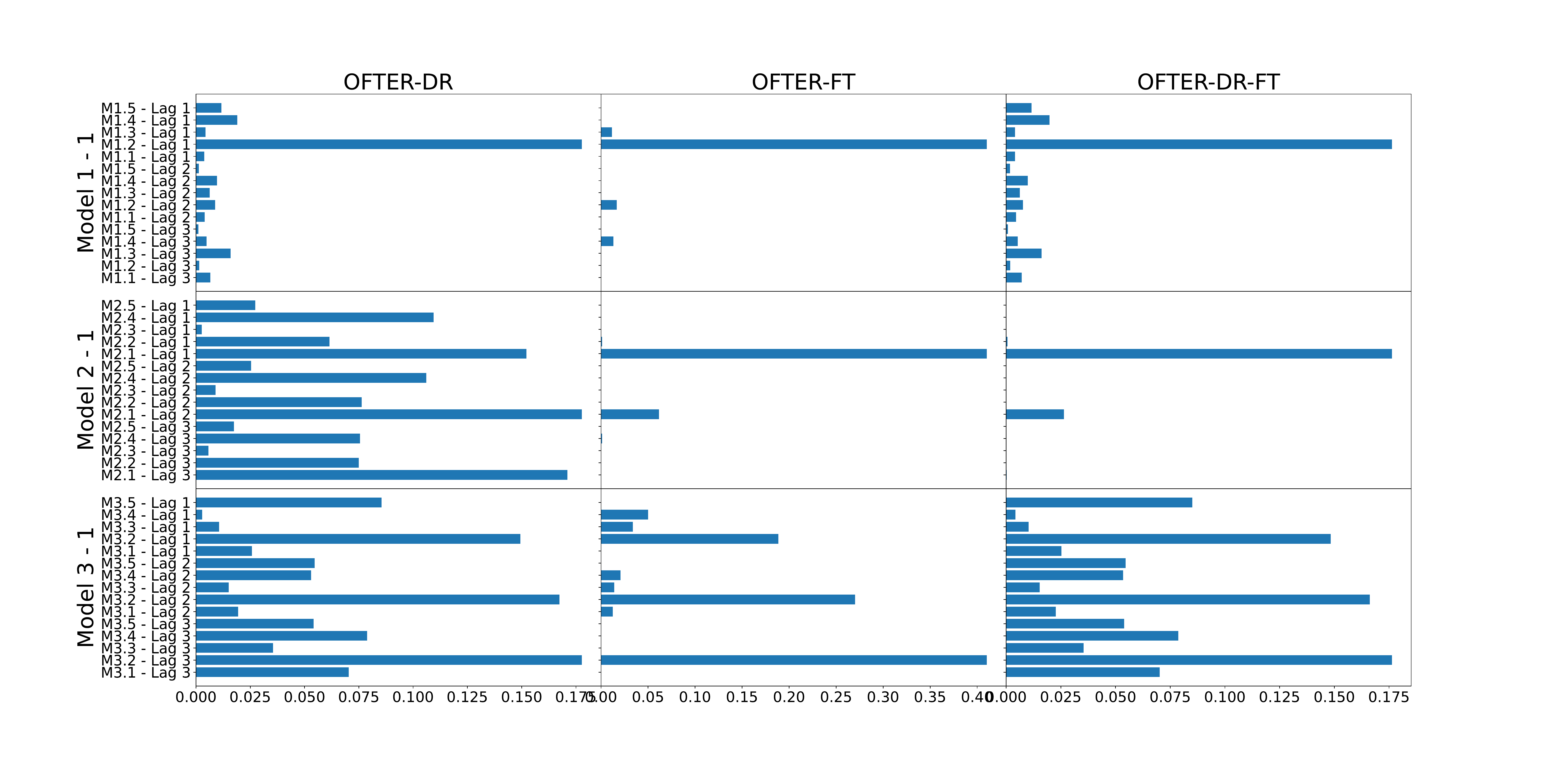}
   \caption{
   }
   \caption{In this 3x3 plot grid, we show the feature importance weights, as obtained in  \Cref{sec:feature_importance}. 
   The first feature of each of M1, M2, and M3 is forecasted in the first, second and third row, respectively.
   In each case, there are 15 features corresponding to three lags of each of the five features. This is repeated for three of {\ofter}'s configurations, corresponding to the three columns of the grid.}
   \label{fig:importance_weights}
\end{figure} 

In the {\ofter} pipeline, one can put forth a natural approach for estimating  the contribution of each feature  to the forecasting task. In particular, we aim to find a proxy for the importance of each of the input features. To this end, consider two points in the initial feature space $x_1$ and $x_2$, and denote by $x_1' = U^T x_1$ and  $x_2' = U^T x_2$ the transformed variables. Denote by $D' = || x_1' - x_2' ||_v$ the distance in the reduced space (used for calculating the nearest neighbors), and let $D_j = x_{1,j}- x_{2,j}$ such that   $||x_1 - x_2||_2 = \sqrt{\sum_j D_j^2}$. 

A metric for the importance of a variable in determining the relative distance of data points in the embedding space can be modeled as the sensitivity of the distance $D'$ on the individual distance $d_j$, i.e $\left| \frac{\partial D'}{\partial D_j}\right|.$ For our goal, it is natural to take the derivative at a point where $D_1 = \dots = D_d = 1$. 
We begin by writing 
\begin{equation} \label{eq:d'2} D'^2 = || x_1' - x_2' ||_{v} = || U^T (x_1 - x_2) ||_{v} = \sum_{j=1}^p v_j \left( \sum_{k=1}^d U_{k,j} (x_1-x_2)_k \right)^2 = \sum_{j=1}^p v_j \left( \sum_{k=1}^d U_{k,j} D_k \right)^2. \end{equation}
We then compute
\begin{equation} \label{importance_prefinal}  \frac{\partial D'^2}{\partial D_k}\Bigr|_{d = e_d}  =  2 \sum_{j=1}^p v_j \left( \sum_{k'=1}^d U_{k',j} D_{k'}  \right) U_{k,j}, 
\end{equation} 
which, using $\odot$ to denote the Hadamard product, can be written as
\begin{equation}
    \nabla D'^2(e_d)
= 2 \left| U  \left[ (U^T e_d) \odot  v \right] \right|. 
\end{equation} 
Keeping in mind \eqref{eq:augment}, we also augment the embedding space with the original features that are highly correlated with the target. Consider the vector $v_2$ containing the weight given to the original feature if used, and zero otherwise, i.e 
\[ v_{2,j} =
\begin{cases}
v_{I^{\textrm{Original}}_k} &\text{$j = I^{\textrm{Original}}_k$} \\
0 &\text{$j \notin I^{\textrm{Original}}$}
\end{cases}.
\]

This now leads to 
\begin{equation} \label{importance_D2} D'^2 = \sum_{j=1}^p v_j \left( \sum_{k=1}^d U_{k,j} D_k \right)^2 + \sum_{l=1}^d v_{2,l} D_l^2,
\end{equation}
and thus the derivative is instead given by
\begin{equation} \label{importance_final}
     \nabla D'^2(e_d) = 2 \left| U  \left[ (U^T e_d) \odot  v \right] + v_2\right|. 
\end{equation}
As an illustration, we consider the task of forecasting the first feature of each of the three synthetic data sets (M1-M3), using {\ofter}-DR, {\ofter}-FT, and {\ofter}-DR-FT. 
In \Cref{fig:importance_weights}, we show a grid of the feature importance of each individual feature. 
As we ignore the unimportant features via the calculation of $v$ in \Cref{eq:v}, {\ofter}-FT attaches a non-zero importance weight only to a subset of the features. On the other hand, {\ofter}-DR assigns a non-zero weight to all of the original features, as they all contribute to the estimation of each principal component. This also holds true for {\ofter}-DR-FT, where dimensionality reduction is also performed. However, in the case of M2 where most of the predictability is present in a single feature (M2.1 - Lag1), {\ofter}-DR-FT assigns a zero importance weight to most of the features, similarly to {\ofter}-FT.  Therefore, for the task of feature importance, we recommend the user to rely on either {\ofter}--FT or {\ofter}-DR-FT.

\subsection{Outlier detection}

\begin{figure}[ht!]  
\centering 
\includegraphics[trim={0cm 0.1cm 0cm 0.4cm},width=1\textwidth]{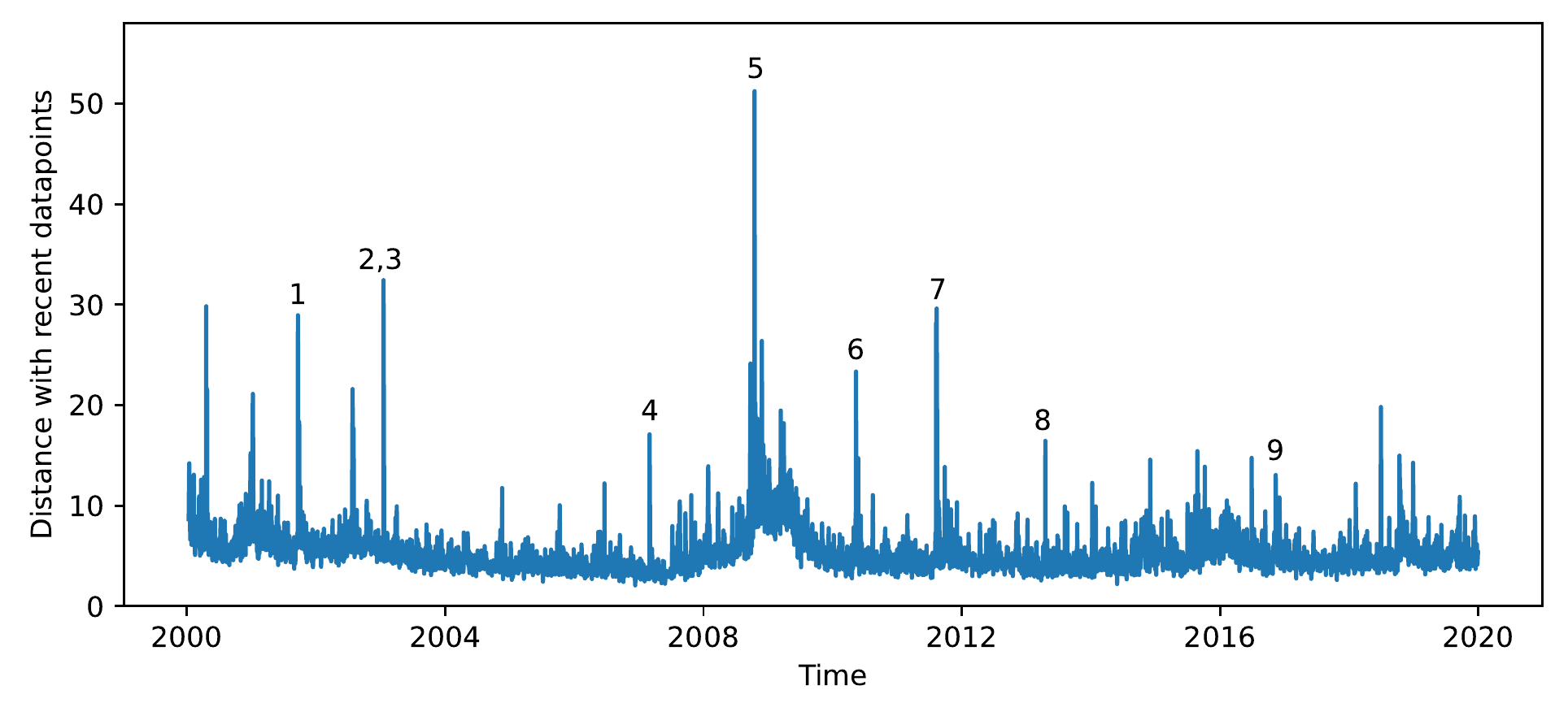}
\caption{Plot showing the minimum distance $d_j^\textrm{min}$ obtained using the suggested outlier detection algorithm for {\ofter}-DR, $L=600$ and $\kappa=5$. All numbered points are identified as outliers. \\
1. 18 September 2001 - Aftermath of the 9/11 Attack \\
2. 24 July 2002 -  Stock market downturn of 2002 ($12^\textrm{th}$ consecutive day of falling prices) \\
3. 29 July 2002 - Stock market downturn of 2002 ($3^\textrm{rd}$ day of rising prices after sharp price rises following (2)) \\
4. 28 February 2007 - Big drop in US and China stocks \\
5. 10 October 2008 - Market Crash across Europe and Asia \\
6. 7 May 2010 - Following the Flash Crash of May 6  \\
7. 4 August 2011 - Preceding the Black Monday (8 August) \\
8. 15 April 2013 - One of the biggest drops in 2013 for DJIA, SPX \\
9. 8 November 2016 - US election}  \label{fig:outlier}
\end{figure} 

A byproduct of the {\ofter} pipeline is that it can facilitate the task of outlier detection, by considering the pairwise distance between time points in the embedding space. More specifically, points whose nearest neighbors are still quite ``far" in the embedding space could be labeled as outliers, in the same spirit as the one-class SVM algorithm \cite{oneclasssvm}. We demonstrate this using the following example. 

Define the minimum distance between the current time point $t$ and the time points in its recent history in the reduced space as 
\begin{equation} \label{eq:outlier_min}
     d_{t}^\textrm{min} = \textrm{min}(\{ d_{t-L,t}, \dots, d_{t-1,t} \}). 
\end{equation}
Comparing $d_{t}^\textrm{min}$ with the distribution of $(d_{t-L}^\textrm{min},\dots,d_{t-1}^\textrm{min})$ may constitute input for any outlier detection algorithm.\footnote{Instead of $d_{t}^\textrm{min}$, one could employ the average distance $\bar{d}_t = \frac{\sum_{j=t-L}^{t-1} d_{j,t}}{t}$.}
Denoting by $q^{\textrm{min}}_{t,1}$ and $q^{\textrm{min}}_{t,3}$ the first and third quantiles of $\{ d^{\textrm{min}}_{t-L}, \dots, d^{\textrm{min}}_{t-1}  \}$,  we classify a time point $t$ as an outlier if $d_{j}^\textrm{min} > q^{\textrm{min}}_{t,3} + \kappa \left( q^{\textrm{min}}_{t,3} - q^{\textrm{min}}_{t,1} \right)$ for some $\kappa > 0$.

In \Cref{fig:outlier}, we plot the minimum distance between each trading day and its $L=600$ preceding days, by using $\kappa = 5$. From this figure, one could infer several different regimes, as there are periods where the levels of $d_\textrm{min}$ are low, and periods where these are high, such as the periods around the 2008 financial crisis. The fact that almost all labeled outliers fall in the immediate vicinity of days significant in the financial history, does indeed indicate that the algorithm can successfully be used to uncover abnormality signals. 

Finally, we remark that selecting a larger value of $L$ will result in a large number of anomalies being highlighted. In practice, if this outlier detection approach is used in a trading application, a smaller $\kappa$ would be desirable, as the outlier warning could possibly be qualitatively analyzed. Note that the above quantile-based outlier detection approach is just a sample algorithm to illustrate the potential of {\ofter}, as the pipeline is amenable to be incorporated into any other outlier detection algorithm.

\section{Conclusion and future work}
\label{sec:conclusion}
In this paper, we have introduced {\ofter}, a complete pipeline for online multivariate time series forecasting. By leveraging kNN and GRNN, both simple well-understood methods with trivial training complexities, {\ofter} manages to outperform a number of baselines, for mid-sized data sets. In particular, an extensive set of numerical experiments demonstrated that {\ofter} performed competitively with Random Forests and GRU in the synthetic data set, and outperformed all other methods in the real-world applications considered, with a running time much lower than those of deep learning methods. The shortcomings due to the curse of dimensionality are overcome via dimensionality reduction and the use of a weighted norm when comparing distances in the embedding space. 

Furthermore, {\ofter} brings  additional advantages, including its flexible nature, with the ability to use or omit many of its components. As a byproduct, {\ofter} can be integrated seamlessly into an 
outlier detection pipeline, in an online manner. Lastly, {\ofter} is naturally equipped to return the user 
a feature importance score for all the input variables.

\textbf{Future work.} 
Future extensions of this work could include the utilization of other machine learning methods in the model selection step of the algorithm, for example, LASSO models with different regularization parameters. One disadvantage of {\ofter} is that it struggles to discover non-linear interaction effects between covariates, even though this is a general issue with models and data sets of this size. One potential extension would be to include a pre-screening component, where a number of potential interaction terms are tested in order to augment the feature space, either prior to or after the embedding step. Additionally, it would be interesting to see the performance of {\ofter} in other mid-sized data sets and assess whether other dimensionality reduction algorithms are able to attain improved performance, relative to the use of PCA (this was not the case for the data sets used in this project). 
Yet another interesting direction concerns an adaptation of {\ofter} for probabilistic forecasting. 
Probabilistic models have recently gained in popularity, being able to provide not only forecasts but also associated uncertainty estimates, which is of crucial importance in certain applications.
Lastly, we note that the OSMC metric we introduced is of independent interest, both as an alternative to the maximal correlation but also as a computationally fast additional step for other simple regression methods. 

\section*{Acknowledgments}
This work was supported by the UK Engineering and Physical Sciences Research Council (EPSRC) grants EP/R513295/1 and EP/V520202/1.

\bibliography{bib_file}
\appendix
\section{A convexifying approach for solving the Secular equation}
In this section, we give a brief description of the method we use to solve the secular equation described in \Cref{sec:update}, which amounts to finding the roots of  
\begin{equation} \label{eq:function_secular}
    f(\kappa) = 1 + \rho \sum_{j=1}^d \frac{z_j^2}{\lambda_j-\kappa}. 
\end{equation}

The main difficulty in directly solving \Cref{eq:function_secular} lies in the multitude of discontinuities in the domain of $f$. An efficient approach to find the $i^{\textrm{th}}$ root of the equation, would be to consider the reparameterizations $\kappa = \lambda_i + \rho \xi$ and $\delta_j = \frac{\lambda_j - \lambda_i}{\rho}$, which leads to the formulation of \Cref{eq:function_secular} as
\begin{equation} \label{eq:function_secular_transformed}
f_i(\xi) = 1 + \sum_{j=1}^d \frac{z_j^2}{\delta_j - \xi}. 
\end{equation}

By assuming that each eigenvalue is distinct (a reasonable assumption when dealing with real data),
we notice that
\begin{equation} \label{eq:delta}
    \delta_1 < \delta_2 < \dots < \delta_i = 0 < \dots< \delta_n.  
\end{equation}
Therefore, we look for solutions in the set $\left( 0, \delta_{i+1} \right)$, for $i < n$, and in $\mathbb{R}_+$ for $i=n$. We can further transform $f_i(x)$ into a convex function, by considering 
$F_i(\gamma) = f_i(\gamma^{-1/2})$, which can be shown to transform $F_i$ into a convex function in the interval $(\frac{1}{\delta_{i+1}},\infty)$ for $i < n$ and $(0,\infty)$ for $i = n$. 

Finally, by using the fact that, for a decreasing convex function, the Newton method\footnote{We remind the reader that the Newton method searches for a solution to $f(x) = 0$, starting from point $x_0$, and then considers the sequence 
\[ x_{n+1} = x_n - \frac{f(x_n)}{f'(x_n)}, \] until a solution is obtained.} monotonically converges to a solution \cite{henrici}, it can be established that, 
if $\gamma^*$ is the solution to $F_i(\gamma) = 0$, then the Newton method converges monotonically to $\gamma^*$ with any starting point in $\left[\frac{1}{\delta_{i+1}^2},\gamma^* \right]$.
For a more detailed explanation and an alternative based on the secant method, we refer the reader to \cite{melman}.

\section{Brief summary of baseline methods}
\label{sec:appendix}
The following methods are used as baselines against {\ofter}. We provide a short description of each one. 

\subsection*{Ordinary Least Squares (OLS)} Ordinary Least Squares (OLS) is by far the most well known regression method, where the target variable $y$ is modeled as a linear combination of the features 
\begin{equation} \label{eq:OLS}
    y_i = \beta_0 + \sum_{j=1}^d \beta_j X_{i,j} + \epsilon_{i}.
\end{equation}  
The coefficient vector $\beta = (\beta_0,\dots,\beta_d)$, is chosen as the one maximizing $\sum \epsilon_{i}^2$, and can be found using the closed form formula 
\begin{equation} \label{eq:ols_beta} \beta = (X^T X)^{-1} X^T y. \end{equation}
Since we include lagged features in the feature matrix, this method is very similar to a Vector Auto-regressive Regression (VAR) model. 

\subsection*{Principal Component Regression} To combat the high correlation present in the features, Principal Component Regression (PCR) first identifies the principal components of $X$, obtaining $\tilde{X}$, and then a number of components $p < d$ is retained. OLS is then used, but on the first $p$ columns of $\tilde{X}$ instead of $X$. Similar to {\ofter}, we use $\delta=0.9$ for choosing the number of components.

\subsection*{Least Absolute Shrinkage and Selection Operator} An alternative to reducing the dimensionality of $X$ before OLS, is to include a regularization term in the objective function \Cref{eq:OLS}. Least Absolute Shrinkage and Selection Operator (LASSO) is a widely popular method, where the modified objective function takes the following form
\begin{equation}
 (y - X \beta)^T (y - X \beta) + \lambda \sum_j | \beta_j |. 
\end{equation}
Note that in this case, there is no closed-form solution, and thus a numerical optimization method is utilized. We choose $\lambda$ using a cross-validation procedure. 

\subsection*{Random Forests}
A decision tree is a simple method inspired by decision analysis, where the visualization of the model resembles a tree consisting of nodes and branches. The initial feature space constitutes the root node. The observations in the root node are partitioned into 'successor nodes', based on a splitting rule chosen so as to optimize some criterion (usually either entropy or Gini Impurity). The process is repeated recursively on each of the successor nodes until some stopping criterion is met. For each new data point, its features are then used to determine its corresponding leaf node. Forecasts can then be obtained by averaging the target value of all observations that fall within each leaf node. This is a non-parametric method, but with a very high variance. 

A Random Forest is built by fitting multiple decision trees. Each decision tree is fitted using only a random sample of observations and a random sample of features. The forecasts of the decision trees are then combined into a single forecast (usually by averaging). Random Forests are based on the idea of bagging aimed at reducing variance and avoiding overfitting. For the benchmark task, we use 200 trees. 

\subsection*{Support Vector Machines (SVM):}
SVM is another popular machine learning model that is motivated by learning theory. In particular, a vector $\beta$ and a value $\beta_0$ are found that minimize
\begin{equation}
    \frac{1}{2} ||\beta||^2, \ \ \ 
    \textrm{s.t}  \ \ \  | (y_i - \beta^T x_i - \beta_0) | \le \epsilon.
\end{equation} 
A prediction for a new point $x'$ is then given by $\beta_0 + \beta^T x'$.

\subsection*{Recurrent Neural Networks:}
Recurrent neural networks (RNN) form a class of neural networks specifically aimed at data sets with temporal structures. This is generally achieved by introducing additional skip connections that pass the output of a past layers to the current one. One issue caused is that gradients have the tendency to vanish or explode.

Long-Short Term Memory is a variant of Recurrent Neural Networks, specifically designed to avoid the issue of exploding or vanishing gradients. This is achieved by introducing a cell state, which contains a summary of the previous cells and is controlled by a number of different gates. Denote by $x_t$ the feature vector of a data point, $c_t$ the input of the cell state, and $h_t$ the output of the module. The formula of the basic LSTM takes the form
\begin{equation} \label{eq:lstm}
\begin{cases}
f_t = \sigma \left(W_f x_t + U_f h_{t-1} + b_f \right), \\
i_t = \sigma \left( W_i x_t + U_i h_{t-1} + b_i \right), \\
o_t = \sigma \left( W_o x_t + U_o h_{t-1} + b_o \right), \\
\hat{c}_t = \textrm{tanh} \left( W_c x_t + U_c h_{t-1} + b_c \right), \\
c_t = f_t \odot c_{t-1} + i_t \odot \hat{c}_t. 
h_t = c_t \odot \textrm{tanh}(c_t), \\
\end{cases}
\end{equation}
where $W_f,U_f,b_f,W_i,U_i,b_i,W_o,U_o,b_o,W_c,U_c$ and $b_c$ are trainable parameters. 

We consider two more variants of LSTM. Bi-LSTM which utilizes two parallel and reverse networks of LSTMs so that information can flow in both directions. We also use Gated Recursion Unit (GRU) which has a similar yet simpler architecture than LSTM. 

For each of these two layers, we use a total of two layers and 300 epochs. 

\section{Robustness analysis} \label{sec:robust}
In this short section, we demonstrate how changing five of the main parameters used in {\ofter}-DR-FT affects its performance on the synthetic data set. We fix five of the main parameters given in {\ofter} as follows: 
$\delta = 0.9$, $L = 800$, $C_\textrm{min} = 0.05$, $C_\textrm{Original} = 0.05$, and $p_{\textrm{ADF}} = 0.05$. By keeping the other four parameters fixed, we vary each of these parameters in turn and note the variation in its performance. In \Cref{tab:robust}, we show the correlation between the target and the forecasts when changing each of the noted parameters to the given value. 

In general, the performance of the algorithm does not significantly depend on the choice of parameters, as forecasting each feature has consistent performance. However there are exceptions; for example, $C_\textrm{min}$ appears to have a significant effect on the performance of the algorithm for M2.3 and M2.4. 

\begin{table}[h]
\centering
\begin{tabular}{@{}ccccccccccc@{}}
   & \multicolumn{2}{c}{\textbf{$\delta$}}        & \multicolumn{2}{c}{\textbf{$p_{\textrm{ADF}}$}}             & \multicolumn{2}{c}{\textbf{$L$}}            & \multicolumn{2}{c}{\textbf{$C_{\textrm{min}}$}}         & \multicolumn{2}{c}{\textbf{$C_{\textrm{original}}$}}    \\
\textbf{Value}                     & \textbf{0.8} & \textbf{0.95}              & \textbf{0.01} & \textbf{0.1}               & \textbf{500} & \textbf{1500}              & \textbf{0.01} & \textbf{0.1}               & \textbf{0.01} & \textbf{0.1}               \\ \midrule
\multicolumn{1}{c|}{\textbf{M1.1}} & 0.470        & \multicolumn{1}{c|}{0.459} & 0.467         & \multicolumn{1}{c|}{0.467} & 0.441        & \multicolumn{1}{c|}{0.477} & 0.464         & \multicolumn{1}{c|}{0.464} & 0.465         & \multicolumn{1}{c|}{0.461} \\
\multicolumn{1}{c|}{\textbf{M1.2}} & 0.519        & \multicolumn{1}{c|}{0.511} & 0.509         & \multicolumn{1}{c|}{0.509} & 0.503        & \multicolumn{1}{c|}{0.521} & 0.509         & \multicolumn{1}{c|}{0.514} & 0.504         & \multicolumn{1}{c|}{0.518} \\
\multicolumn{1}{c|}{\textbf{M1.3}} & 0.077        & \multicolumn{1}{c|}{0.071} & 0.085         & \multicolumn{1}{c|}{0.085} & 0.085        & \multicolumn{1}{c|}{0.113} & 0.119         & \multicolumn{1}{c|}{0.116} & 0.085         & \multicolumn{1}{c|}{0.117} \\
\multicolumn{1}{c|}{\textbf{M1.4}} & 0.554        & \multicolumn{1}{c|}{0.558} & 0.552         & \multicolumn{1}{c|}{0.552} & 0.555        & \multicolumn{1}{c|}{0.556} & 0.553         & \multicolumn{1}{c|}{0.556} & 0.545         & \multicolumn{1}{c|}{0.559} \\
\multicolumn{1}{c|}{\textbf{M1.5}} & 0.094        & \multicolumn{1}{c|}{0.107} & 0.101         & \multicolumn{1}{c|}{0.101} & 0.147        & \multicolumn{1}{c|}{0.120} & 0.102         & \multicolumn{1}{c|}{0.110} & 0.101         & \multicolumn{1}{c|}{0.113} \\
\multicolumn{1}{c|}{\textbf{M2.1}} & 0.599        & \multicolumn{1}{c|}{0.623} & 0.599         & \multicolumn{1}{c|}{0.599} & 0.596        & \multicolumn{1}{c|}{0.603} & 0.675         & \multicolumn{1}{c|}{0.600} & 0.599         & \multicolumn{1}{c|}{0.60}  \\
\multicolumn{1}{c|}{\textbf{M2.2}} & 0.535        & \multicolumn{1}{c|}{0.535} & 0.535         & \multicolumn{1}{c|}{0.535} & 0.534        & \multicolumn{1}{c|}{0.539} & 0.897         & \multicolumn{1}{c|}{0.535} & 0.536         & \multicolumn{1}{c|}{0.536} \\
\multicolumn{1}{c|}{\textbf{M2.3}} & 0.864        & \multicolumn{1}{c|}{0.799} & 0.815         & \multicolumn{1}{c|}{0.815} & 0.776        & \multicolumn{1}{c|}{0.85}  & 0.759         & \multicolumn{1}{c|}{0.401} & 0.804         & \multicolumn{1}{c|}{0.817} \\
\multicolumn{1}{c|}{\textbf{M2.4}} & 0.962        & \multicolumn{1}{c|}{0.964} & 0.963         & \multicolumn{1}{c|}{0.963} & 0.963        & \multicolumn{1}{c|}{0.963} & 0.962         & \multicolumn{1}{c|}{0.963} & 0.963         & \multicolumn{1}{c|}{0.963} \\
\multicolumn{1}{c|}{\textbf{M2.5}} & 0.679        & \multicolumn{1}{c|}{0.681} & 0.681         & \multicolumn{1}{c|}{0.681} & 0.678        & \multicolumn{1}{c|}{0.682} & 0.681         & \multicolumn{1}{c|}{0.678} & 0.680         & \multicolumn{1}{c|}{0.679} \\
\multicolumn{1}{c|}{\textbf{M3.1}} & 0.758        & \multicolumn{1}{c|}{0.758} & 0.758         & \multicolumn{1}{c|}{0.758} & 0.751        & \multicolumn{1}{c|}{0.762} & 0.758         & \multicolumn{1}{c|}{0.758} & 0.758         & \multicolumn{1}{c|}{0.757} \\
\multicolumn{1}{c|}{\textbf{M3.2}} & 0.852        & \multicolumn{1}{c|}{0.851} & 0.851         & \multicolumn{1}{c|}{0.851} & 0.85         & \multicolumn{1}{c|}{0.852} & 0.851         & \multicolumn{1}{c|}{0.850} & 0.849         & \multicolumn{1}{c|}{0.850} \\
\multicolumn{1}{c|}{\textbf{M3.3}} & 0.669        & \multicolumn{1}{c|}{0.664} & 0.665         & \multicolumn{1}{c|}{0.665} & 0.662        & \multicolumn{1}{c|}{0.668} & 0.665         & \multicolumn{1}{c|}{0.666} & 0.664         & \multicolumn{1}{c|}{0.667} \\
\multicolumn{1}{c|}{\textbf{M3.4}} & 0.902        & \multicolumn{1}{c|}{0.903} & 0.903         & \multicolumn{1}{c|}{0.903} & 0.902        & \multicolumn{1}{c|}{0.905} & 0.903         & \multicolumn{1}{c|}{0.902} & 0.903         & \multicolumn{1}{c|}{0.902} \\
\multicolumn{1}{c|}{\textbf{M3.5}} & 0.826        & \multicolumn{1}{c|}{0.827} & 0.827         & \multicolumn{1}{c|}{0.827} & 0.825        & \multicolumn{1}{c|}{0.828} & 0.827         & \multicolumn{1}{c|}{0.826} & 0.825         & \multicolumn{1}{c|}{0.825}
\end{tabular} 
\caption{Robustness analysis of {\ofter}. The table displays the correlation between the target and forecasts for each of the 15 features of the synthetic data set, by varying a subset of five parameters of the algorithm.} 
\label{tab:robust}
\end{table} 
\end{document}